\documentclass[11pt]{article}

\usepackage[preprint]{acl}
\makeatletter\newif\ifanonymous\ifacl@anonymize\anonymoustrue\fi\makeatother

\usepackage{times}
\usepackage{latexsym}
\usepackage[T1]{fontenc}
\usepackage[utf8]{inputenc}
\usepackage{microtype}
\usepackage{inconsolata}
\usepackage{graphicx}

\usepackage{url}
\usepackage{booktabs}
\usepackage{amsfonts}
\usepackage{amsmath}
\usepackage{nicefrac}
\usepackage{xcolor}
\usepackage{tabularx}
\usepackage{array}
\usepackage{colortbl}
\usepackage{subcaption}
\usepackage[above]{placeins}  
\usepackage{float}       
\usepackage{makecell}
\usepackage{multirow}
\usepackage{tikz}

\definecolor{ShapeObserved}{HTML}{EAF3FF}
\definecolor{ShapeHidden}{HTML}{FFF1E6}
\definecolor{ShapeDerived}{HTML}{F2F2F2}

\newcolumntype{Y}{>{\raggedright\arraybackslash}X}

\usepackage[normalem]{ulem}

\usepackage[most]{tcolorbox}
\usepackage{needspace}
\usepackage{newfloat}
\DeclareFloatingEnvironment[fileext=lop,name=Prompt,placement=p]{prompt}
\definecolor{promptbg}{gray}{0.95}
\definecolor{AgentBar}{HTML}{3C6FB6}
\definecolor{UserBar}{HTML}{C8692A}
\definecolor{PlannerBar}{HTML}{3F8F5F}
\definecolor{PromptBody}{HTML}{FAFBFC}
\newtcolorbox{prompttitle}[1][AgentBar]{enhanced,colback=#1,colframe=#1,boxrule=0pt,arc=1.5pt,sharp corners=south,left=6pt,right=6pt,top=2pt,bottom=2pt,before skip=8pt,after skip=0pt,fontupper=\small\bfseries\sffamily,coltext=white}
\newtcolorbox{promptpar}[1][AgentBar]{enhanced,colback=PromptBody,colframe=#1,boxrule=0pt,leftrule=0.6pt,rightrule=0.6pt,arc=0pt,left=6pt,right=6pt,top=4pt,bottom=4pt,before skip=0pt,after skip=0pt,fontupper=\small\ttfamily\frenchspacing\raggedright,before upper={\setlength{\parskip}{4pt}\setlength{\parindent}{0pt}}}
\newtcolorbox{promptlast}[1][AgentBar]{enhanced,colback=PromptBody,colframe=#1,boxrule=0pt,leftrule=0.6pt,rightrule=0.6pt,bottomrule=0.6pt,arc=1.5pt,sharp corners=north,left=6pt,right=6pt,top=4pt,bottom=5pt,before skip=0pt,after skip=8pt,fontupper=\small\ttfamily\frenchspacing\raggedright,before upper={\setlength{\parskip}{4pt}\setlength{\parindent}{0pt}}}
\newtcolorbox{promptsec}[1][AgentBar]{enhanced,colback=PromptBody,colframe=#1,boxrule=0pt,leftrule=0.6pt,rightrule=0.6pt,arc=0pt,left=6pt,right=6pt,top=5pt,bottom=4pt,before skip=0pt,after skip=0pt,fontupper=\small\ttfamily\frenchspacing\raggedright,before upper={\setlength{\parskip}{4pt}\setlength{\parindent}{0pt}},overlay={\draw[#1!35,line width=0.5pt] ([xshift=0.6pt]frame.north west) -- ([xshift=-0.6pt]frame.north east);}}
\newtcolorbox{promptseclast}[1][AgentBar]{enhanced,colback=PromptBody,colframe=#1,boxrule=0pt,leftrule=0.6pt,rightrule=0.6pt,bottomrule=0.6pt,arc=1.5pt,sharp corners=north,left=6pt,right=6pt,top=5pt,bottom=5pt,before skip=0pt,after skip=8pt,fontupper=\small\ttfamily\frenchspacing\raggedright,before upper={\setlength{\parskip}{4pt}\setlength{\parindent}{0pt}},overlay={\draw[#1!35,line width=0.5pt] ([xshift=0.6pt]frame.north west) -- ([xshift=-0.6pt]frame.north east);}}

\newcommand{\promptph}[1]{\textcolor{gray}{\rmfamily\itshape [#1]}}
\newtcolorbox{turnbox}[2]{enhanced,colback=#1,colframe=#2,boxrule=0pt,leftrule=2.5pt,arc=1pt,left=5pt,right=5pt,top=3pt,bottom=3pt,before skip=5pt,after skip=5pt,fontupper=\small\ttfamily\frenchspacing\raggedright}
\newcommand{\agentturn}[2]{\begin{turnbox}{ShapeObserved}{AgentRule}\textbf{\sffamily\textcolor{AgentRule}{#1}}\quad #2\end{turnbox}}
\newcommand{\userturn}[2]{\begin{turnbox}{ShapeHidden}{UserRule}\textbf{\sffamily\textcolor{UserRule}{#1}}\quad #2\end{turnbox}}
\definecolor{AgentRule}{HTML}{3C6FB6}
\definecolor{UserRule}{HTML}{C8692A}

\usepackage{xspace}
\newcommand{\taskname}{ReSHAPE\xspace}
\newcommand{\benchname}{ReSHAPE Travel\xspace}
\newcommand{\pipelinename}{ReSHAPE Planter\xspace}

\usepackage{color-edits}
\addauthor{sw}{purple}
\addauthor{swc}{blue}

\title{How Should Planning Agents ReSHAPE Their Search Space?}

\author{
  Yilin Zhang$^{1}$ \quad
  Xinran Zhao$^{1}$ \quad
  Tong Chen$^{2}$ \quad
  Ziming Yuan$^{1}$ \quad
  Tongshuang Wu$^{1}$ \\
  $^{1}$Carnegie Mellon University \quad
  $^{2}$University of Washington \\
  Correspondence: \texttt{\{jasonzh3, sherryw\}@cs.cmu.edu}}

\begin{document}

\maketitle

\begin{abstract}
Planning agents and benchmarks treat a user's initial specification as fixed.
Yet the requests users write exclude plans they would gladly take, since they rarely discover the possible trade-offs
they would accept until seeing the options.
Agents already revise requests through interaction, but whether this helps planning is unknown.
We formalize this step as the \taskname task, which judges a revision by whether its search space matches the
acceptable plans and a planner finds one there.
Using \pipelinename, which builds instances from planning benchmarks with a human-validated simulated user, we analyze
5,800 agent-revised specifications in travel and shopping.
We find that reshaping matters: one correct trade-off makes the same planner reach an acceptable plan up to
4.6$\times$ as often, and search space quality predicts what the planner finds.
Current agents, however, admit at most 32\% of the acceptable plans, where that trade-off alone admits over 52\%,
because they ask about preferences instead of showing plans; in travel, showing plans raises $\mathrm{F}_1$ by 9.7 points over asking, with two model families.
Planning agents should thus treat a request as a hypothesis to test against the user's reactions to concrete plans to
recognize user trade-offs.
\end{abstract}

\begin{figure*}[t!]
  \centering
  \includegraphics[width=\textwidth]{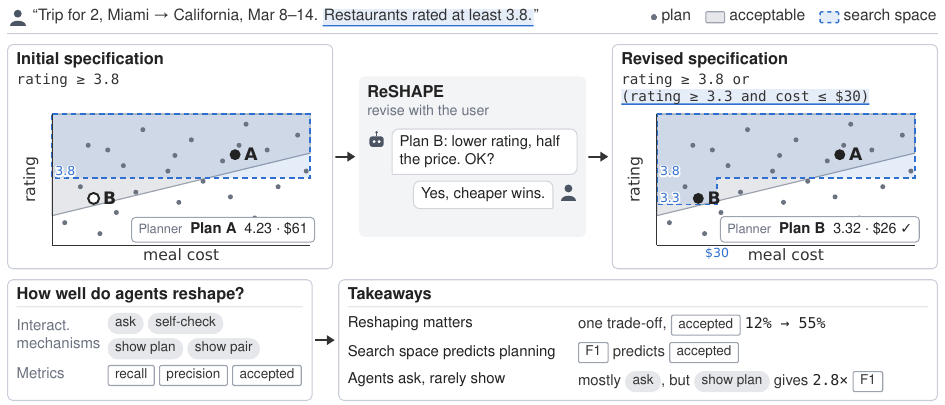}
  \caption{\taskname scores the revision step before planning. \textbf{Top left:} the request's specification excludes Plan B, which the user would accept. \textbf{Top:} an agent revises the specification with the user, and the planner, searching the revised space, finds Plan B. \textbf{Bottom left:} the kinds of interaction found in agents' logs, which we compare under a fixed feedback budget, and the metrics that score a revision: the recall and precision of its search space against the acceptable plans, and whether a fixed planner delivers an accepted plan. \textbf{Bottom right:} our main findings.}
  \label{fig:teaser}
\end{figure*}

\section{Introduction}
\label{sec:introduction}

Language-model agents are increasingly asked to turn a natural-language request into a concrete plan with many parts, such as a multi-day trip.
The request is first turned into a specification, the conditions a plan must meet~\citep{cook-2024-tick,hao-2025-formal-verification-planning,shao-2025-personal-travel-solver}, and a planner then searches for a plan that satisfies it. Planning benchmarks follow the same pipeline: they evaluate whether the final plan satisfies the specification the user provides~\citep{xie-2024-travelplanner,zheng-2024-natural-plan,xiao-2024-flowbench}.
\textbf{A plan is therefore only as good as the specification it is planned against}: a planner cannot return a plan that the specification excludes, and it struggles to find a good plan in one that admits too much.

Yet a specification built from the user's request is a poor guide to the plans they would accept, because people discover what they would accept only as they encounter concrete options~\citep{bettman-1998-constructive-consumer-choice,pirolli-2005-sensemaking-process,kim-2026-discoverllm}.
A traveler who asks for restaurants rated at least 3.8 may gladly accept lower-rated ones that cost less than half as much, but cannot say in advance how much lower is acceptable, since that depends on the restaurants actually on offer (Figure~\ref{fig:teaser}, top left).
Such trade-offs, in which a strength on one aspect of a plan compensates for a weakness on another~\citep{keeney-1976-decisions-multiple-objectives}, rarely make it into a request, so the initial specification excludes plans the user would accept.

Existing agents already try to close this gap through interaction. They ask clarifying questions before planning~\citep{qian-2024-in3,zhang-2024-ask-before-plan}, elicit preferences the user left unstated~\citep{qian-2025-userbench}, or replan as new constraints arrive~\citep{oh-2025-flextravelplanner,liu-2026-adaplanbench}. The same step appears in commercial agents: OpenAI's and Google's Deep Research agents confirm the request or propose a research plan for the user to revise~\citep{openai-2025-deep-research-api,google-2024-gemini-deep-research}.
These settings, however, judge the plan or choice finally delivered, which conflates the revised request with the planner that acts on it~\citep{shen-2025-completion-collaboration}, so whether a revision recovers the plans the request excluded goes unmeasured.

\looseness=-1 We want to know whether this revision step actually pays off for planning, what makes a revised specification good for a planner, and what an agent has to do to produce one.
To answer these questions, we formalize this step as the \textbf{\taskname} task (\S\ref{sec:reshape}). Before the planner runs, an agent interacts with the user to revise the initial specification into one that includes the trade-offs the user would accept. The planner then searches only the plans the revised specification admits (Figure~\ref{fig:teaser}, top).
The user is simulated with hidden preferences, so whether a plan is acceptable can be computed exactly, while the agent, as in practice, learns those preferences only through a limited budget of questions and shown plans.
In our user study, the simulated user's acceptance decisions agree with human participants' as often as participants agree with one another (\S\ref{sec:human-validation}).
A revision can then be scored at two levels (Figure~\ref{fig:teaser}, bottom left): upstream, by how its \emph{search space} overlaps the \emph{acceptable plans}, with \emph{recall} for how many of the acceptable plans it admits and \emph{precision} for how many of the admitted plans are acceptable; and downstream, by whether a fixed planner given the revision delivers an acceptable plan.
Such scoring needs instances in which revision is both necessary and possible; we thus build \textbf{\pipelinename}, a pipeline that plants in each instance of a planning benchmark a trade-off that the initial specification excludes and a reference specification adds.

As a proof of concept, we apply \pipelinename to TravelPlanner~\citep{xie-2024-travelplanner} and to the shopping tasks of DeepPlanning~\citep{zhang-2026-deepplanning} to build \benchname and ReSHAPE Shopping, on which we analyze the revisions of 14 and 8 off-the-shelf agent systems, respectively; on \benchname we also compare four interaction protocols, each limited to one kind of interaction found in these agents' logs and run with the same model and feedback budget.
Our analysis then yields three findings about the step itself (\S\ref{sec:usefulness}; Figure~\ref{fig:teaser}, bottom right).
First, \textbf{reshaping the specification matters}: with the reference specification, which adds a single trade-off, the same planner finds a plan the user would accept up to 4.6$\times$ as often as with the initial one.
Second, \textbf{search space quality predicts the planning outcome}: in both domains, the $\mathrm{F}_1$ of recall and precision tracks whether the planner delivers an acceptable plan, so a revision can be judged before any planner runs.
Third, \textbf{a good revision first admits more plans and then stays selective.}
Early on only recall matters: the initial specification admits almost no acceptable plans, so each one a revision lets back in raises the chance that the planner finds it.
Once the planner has enough to choose from, more recall stops helping and precision decides, because a search space full of plans the user would reject sends the planner to the wrong ones.

\looseness=-1 \textbf{Current agents prioritize precision but cannot recover recall} (\S\ref{sec:agents}): they admit at most 32\% of the acceptable plans, against over 52\% for the reference specification, because they spend their effort asking about preferences that users rarely know before seeing the options.
The systems rank nearly the same on ReSHAPE Shopping in precision and in what the planner finds (Table~\ref{tab:cross-domain}), so this shortfall belongs to the agents rather than to a specific domain.
\textbf{Showing the user plans outside the current specification helps mitigate the imbalance}: with the same model and feedback budget, it raises $\mathrm{F}_1$ by 9.7 points over asking (95\% CI [5.0, 15.0]) with Gemini 3.6 Flash and by 11.2 points with Claude Sonnet 4.6.
Current agents thus treat the request as a fact to clarify, when it should be a hypothesis to test against the user's reactions to concrete plans.

In sum, we (1) make the revision step before planning measurable on its own, apart from the planner that follows it; (2) build \pipelinename and, with it, \benchname and ReSHAPE Shopping; and (3) show where current agents fall short in this step and that showing plans, not asking, is what recovers the plans a request excludes.

\section{Related Work}
\label{sec:related_works}

\paragraph{Stated preferences omit trade-offs that concrete options reveal.}
Stated requirements typically take the form of separate cutoffs that reject an alternative as soon as one of them fails
\citep{einhorn-1970-noncompensatory,payne-1993-adaptive-decision-maker}, whereas judgments of complete alternatives are
compensatory \citep{keeney-1976-decisions-multiple-objectives}.
Decision analysis therefore elicits trade-offs by having people compare specific alternatives rather than state them in
the abstract \citep{keeney-1976-decisions-multiple-objectives}, and example-critiquing recommenders refine a user's
preference model from reactions to displayed items \citep{chen-2012-critiquing,viappiani-2006-example-critiquing}.
We test whether this practice carries over to planning agents, comparing protocols that show the user concrete plans
with protocols that ask about preferences under the same feedback budget.

\paragraph{Interactive planning revises requests but judges the delivered plan.}
Interactive planning settings let agents ask clarifying questions, elicit hidden preferences, or revise a plan after
feedback \citep{qian-2024-in3,zhang-2024-ask-before-plan,oh-2025-flextravelplanner,qian-2025-userbench,liu-2026-adaplanbench},
so they already revise the request in effect, but they score the final plan or choice, and few show the user plans or involve trade-offs
(Appendix~\ref{app:related-settings}).
DiscoverLLM \citep{kim-2026-discoverllm} simulates users whose intents concretize as the model
surfaces options, and finds that asking fails when users do not yet know what they want; like the settings above, it judges the final artifact.
We instead hold the user's acceptance of every plan fixed, which lets us score a revised specification itself over all
legal plans and relate this score to what a planner finds within it.

\paragraph{Existing work revises specifications, but not by the plans they admit.}
When constraints conflict, relaxation methods choose small or low-priority corrections that restore feasibility
\citep{narodytska-2018-correction-sets,eifler-2022-soft-goal-conflicts,li-2023-prioritized-preferences}; constraint
acquisition learns intended constraints from queries, examples, or natural-language feedback
\citep{belaid-2022-geqca,tsouros-2023-growacq,tsouros-2025-gencon,mechqrane-2024-llmacq}; and work on objective
misspecification treats a stated goal as a proxy that interaction can correct
\citep{he-2022-assisted-reward-design,mechergui-2024-goal-alignment,mechergui-2024-expectation-alignment}.
All judge a revision by whether it restores feasibility or recovers a single target specification.
Closest is GATE, which tests whether elicited preferences predict a user's decisions on held-out examples
\citep{li-2025-eliciting-human-preferences}; we carry this idea to planning, judging a revision by how well its search
space matches the plans the user would accept (\S\ref{sec:evaluation-metrics}) and testing which way of scoring this
match best predicts what a planner finds (\S\ref{sec:usefulness}).

\section{Making Specification Revision Measurable}
\label{sec:reshape}
\label{sec:task_definition}
\label{sec:benchmark}

\taskname inserts a revision step before the planner: given the same input as planning, an agent interacts with the user and returns a revised specification instead of a plan.
This section defines the step (\S\ref{sec:task-formulation}), the simulated user that makes acceptance computable (\S\ref{sec:simulated-user}), and how we measure a revision (\S\ref{sec:evaluation-metrics}); \S\ref{sec:experimental-setup} introduces \pipelinename, which builds instances from planning benchmarks.

\subsection{\taskname: Revision before Planning}
\label{sec:task-formulation}

\paragraph{Background: plans and specifications.}
A planning task starts from a user's request and the available options, grouped into categories such as flights, hotels, and restaurants.
Current pipelines turn the request into a \emph{specification} that lists conditions on plan attributes, such as a minimum restaurant rating or a maximum meal cost \citep{hao-2025-formal-verification-planning,shao-2025-personal-travel-solver}, and a planner searches for a complete plan satisfying every condition.

Intuitively, a request carries two kinds of content.
Part of it is fixed, such as the destination, the dates, and the size of the group; we call this part the query $q$, and the complete plans that respect it the \emph{legal plans} $\mathcal{P}$.
The rest consists of \emph{preferences}, such as the rating: stated as if they were equally strict rules, but most of them the user would readily trade for something they value more. We collect them in the \emph{initial specification} $C_0$, writing $c_i$ for a condition on attribute $i$.
A planner treats both parts alike, so it considers only the legal plans that also satisfy the specification, the \emph{search space} $\mathcal{S}(C)=\{p\in\mathcal{P}\mid p\text{ satisfies }C\}$.

\paragraph{The \taskname task.}
\looseness=-1 We define \taskname as the step before the planner.
It takes the same input as planning: the query, its options, and the initial specification $C_0$, and instead of a plan it returns a revised specification $C'$ that the planner follows.

The plans the user would actually accept form another subset of the legal plans, the \emph{acceptable plans} $\mathcal{A}$, which does not necessarily coincide with the search space: the request was written before the user saw any concrete option, and users discover which plans they would accept only as they see them (\S\ref{sec:introduction}, Figure~\ref{fig:teaser}).
The task is to return a $C'$ whose search space matches $\mathcal{A}$ as closely as possible, including plans the user never saw.

\paragraph{Ground truth $\mathcal{A}$ and the expected $C'$.}
To make revision measurable, we need a ground truth: for every legal plan, whether the user would accept it.
We define $\mathcal{A}$ from how people actually decide, with three literature-grounded properties.
First, a strength on one attribute can offset a weakness on another \citep{keeney-1976-decisions-multiple-objectives}, so no fixed threshold per attribute separates the plans the user accepts from those they reject, and $\mathcal{A}$ generally cannot be written as a conjunction of separate conditions (Figure~\ref{fig:teaser}).
Second, the user accepts nothing worse than what the request already secures, up to a small tolerance below which plans are indistinguishable \citep{luce-1956-semiorders}, so acceptance measures what a revision adds beyond $C_0$.
Third, a few conditions would be deal-breakers that nothing offsets \citep{einhorn-1970-noncompensatory}; they are stated in $C_0$ like any other preference, and the agent is not told which they are.
We realize $\mathcal{A}$ through a simulated user that fixes these judgments for every plan yet reveals them only through interaction (\S\ref{sec:simulated-user}).

Since $\mathcal{A}$ is compensatory, a revised specification has to be able to express a trade-off, which a plain conjunction cannot. Beyond loosening or tightening the conditions in $C_0$, $C'$ may use the \emph{trade-off form} $c_i \lor (c_i^{-}\land c_j)$: it also admits plans that violate $c_i$ but satisfy a strictly relaxed version $c_i^{-}$, provided that a compensating condition $c_j$ on another attribute holds (Appendix~\ref{app:constraint-language}). For example, a trade-off item can admit restaurants rated at least 3.8, or at least 3.3 if meals average under \$30.

\subsection{Simulated User and Interaction}
\label{sec:simulated-user}

\paragraph{Hidden profile.}
Each instance holds a hidden profile that decides every question or shown plan, so that the user's judgments never reach the agent directly. The agent only sees that decision verbalized by a language model, which never sees the profile and answers as a user would, without numbers or thresholds (Appendix~\ref{app:simulated-user}).

\paragraph{Utility and acceptance.}
\looseness=-1 The profile realizes the three properties of \S\ref{sec:task-formulation} through a utility $U(p)$ over complete plans.
For compensation, $U$ is a multi-attribute utility \citep{keeney-1976-decisions-multiple-objectives} that rewards doing better than the user expects on an attribute and penalizes falling short, so a gain on one attribute can make up for a loss on another (Appendix~\ref{app:hidden-user-model}).
For reference dependence, we take the best plan the initial specification admits, $p_0^{*}$, as the reference point, and allow a tolerance of $\delta=0.05$ below it, which gives the acceptance threshold $\tau=U(p_0^{*})-\delta$; $C_0$ therefore always admits an acceptable plan, and revision is needed to reach the rest.
For the deal-breaker, we mark one condition in $C_0$ as non-negotiable: a plan that violates it is rejected whatever its utility (Appendix~\ref{app:tolerance}).
The acceptable plans are $\mathcal{A}=\{p\in\mathcal{P}\mid p\text{ meets the deal-breaker and }U(p)\ge\tau\}$.

\paragraph{Interaction.}
Since the profile is hidden, the only way for an agent to perform \taskname is to converse with the user: it can ask a preference question or show a plan and hear the user's judgment of it (Figure~\ref{fig:teaser}).
To stay practical, we guard the conversation with a feedback budget $B$, in line with real users' limited willingness to iterate \citep{shen-2025-completion-collaboration}: each answer and each judgment costs one unit, while inspecting options and checking candidate plans against the specification cost nothing (Appendix~\ref{app:task-interface}).

\subsection{Measuring a Revision}
\label{sec:evaluation-metrics}

\paragraph{Intrinsic: search space quality.}
We score $\mathcal{S}(C')$ against $\mathcal{A}$ as a binary classification over the legal plans, where a plan is a true positive if $C'$ admits it and the user accepts it.
Plans do not count equally, though: the user is nearly indifferent about a plan whose utility sits near the threshold $\tau$, and cares a great deal about one far above or far below it.
We therefore measure \textbf{recall}, how much of what the user would accept the revision admits, and \textbf{precision}, how much of what it admits the user would accept, in plan weight rather than plan count.
A plan's weight is its distance from the threshold, $w(p)=\lvert U(p)-\tau\rvert$. Writing $w(X)$ for the total weight of a set of plans, weighted recall is $R_w=w(\mathcal{S}(C')\cap\mathcal{A})/w(\mathcal{A})$ and weighted precision is $P_w=w(\mathcal{S}(C')\cap\mathcal{A})/w(\mathcal{S}(C'))$. We report both and their harmonic mean, the \textbf{weighted $\mathrm{F}_1$} ($\mathrm{F}_{1,w}$), on a 0--100 scale.
All sums are exact over every legal plan (Appendix~\ref{app:region-computation}).

\paragraph{Extrinsic: planning outcome.}
\looseness=-1 A specification ultimately matters through the plans built within it, so we also give $C'$ to a fixed planner, which sees the instance and the specification but not the user, and ask it to deliver a legal plan within $\mathcal{S}(C')$.
Our main planning measure is the \textbf{acceptance rate} (Acc), the share of instances whose delivered plan lies in $\mathcal{A}$, counting an instance without a delivered plan as not accepted.
We also report the \textbf{delivery rate} (Deliv), the share of instances with a delivered plan, and give the utility of
delivered plans in Appendix~\ref{app:region-change}.

\section{Experimental Setup}
\label{sec:experiments}
\label{sec:main-experiment}
\label{sec:experimental-setup}

\subsection{Benchmarks}
\label{sec:benchmarks}

\begin{table*}[t!]
\begin{minipage}[t]{\columnwidth}\vspace{0pt}
\centering
\footnotesize
\setlength{\tabcolsep}{1.0pt}
\renewcommand{\arraystretch}{1.3}
\newcolumntype{N}{>{\centering\arraybackslash}p{24.5pt}}   
\begin{tabular}{@{}cl NNN NN@{}}
\toprule
\multicolumn{2}{@{}l}{\textit{\benchname}} & \multicolumn{3}{c}{Specification} & \multicolumn{2}{c}{Accepted (\%)} \\
\cmidrule(lr){3-5}\cmidrule(l){6-7}
& Model & $\mathrm{F}_{1,w}$ & $P_w$ & $R_w$ & Flash & Luna \\
\midrule
\multicolumn{2}{@{}l}{$C_{\mathrm{ref}}$ (reference)} & 21.5 & 26.5 & 53.8 & 55 & 47 \\
\midrule
 & Kimi K3 & \cellcolor[HTML]{b5dec1}12.2 & \cellcolor[HTML]{98d1aa}19.2 & \cellcolor[HTML]{b5dec1}31.5 & \cellcolor[HTML]{98d1aa}36 & \cellcolor[HTML]{7bc492}44 \\
 & Claude Sonnet 4.6 & \cellcolor[HTML]{b5dec1}9.5 & \cellcolor[HTML]{7bc492}23.5 & \cellcolor[HTML]{b5dec1}22.6 & \cellcolor[HTML]{7bc492}\textbf{47} & \cellcolor[HTML]{7bc492}42 \\
 & Gemini 3.6 Flash & \cellcolor[HTML]{b5dec1}8.7 & \cellcolor[HTML]{7bc492}24.5 & \cellcolor[HTML]{d2ebd9}14.7 & \cellcolor[HTML]{98d1aa}34 & \cellcolor[HTML]{98d1aa}36 \\
 & Gemini 3.1 Pro & \cellcolor[HTML]{d2ebd9}4.4 & \cellcolor[HTML]{98d1aa}16.2 & \cellcolor[HTML]{eaf6ee}5.4 & \cellcolor[HTML]{d2ebd9}14 & \cellcolor[HTML]{d2ebd9}14 \\
\multirow{-5}{*}{{\scriptsize\bfseries\rotatebox[origin=c]{90}{OpenHands}}} & gpt-oss-120b & \cellcolor[HTML]{eaf6ee}2.1 & \cellcolor[HTML]{d2ebd9}6.0 & \cellcolor[HTML]{d2ebd9}13.1 & \cellcolor[HTML]{eaf6ee}10 & \cellcolor[HTML]{d2ebd9}10 \\
\midrule
 & Kimi K3 & \cellcolor[HTML]{98d1aa}\textbf{14.0} & \cellcolor[HTML]{7bc492}25.9 & \cellcolor[HTML]{b5dec1}\textbf{31.6} & \cellcolor[HTML]{7bc492}44 & \cellcolor[HTML]{7bc492}44 \\
 & Claude Sonnet 4.6 & \cellcolor[HTML]{b5dec1}12.1 & \cellcolor[HTML]{7bc492}22.3 & \cellcolor[HTML]{b5dec1}22.5 & \cellcolor[HTML]{98d1aa}43 & \cellcolor[HTML]{98d1aa}33 \\
 & Gemini 3.6 Flash & \cellcolor[HTML]{d2ebd9}8.2 & \cellcolor[HTML]{98d1aa}18.6 & \cellcolor[HTML]{d2ebd9}18.4 & \cellcolor[HTML]{b5dec1}31 & \cellcolor[HTML]{b5dec1}22 \\
 & Gemini 3.1 Pro & \cellcolor[HTML]{d2ebd9}7.3 & \cellcolor[HTML]{98d1aa}20.8 & \cellcolor[HTML]{d2ebd9}13.2 & \cellcolor[HTML]{b5dec1}25 & \cellcolor[HTML]{98d1aa}30 \\
\multirow{-5}{*}{{\scriptsize\bfseries\rotatebox[origin=c]{90}{mini-SWE-agent}}} & gpt-oss-120b & \cellcolor[HTML]{eaf6ee}2.9 & \cellcolor[HTML]{eaf6ee}4.8 & \cellcolor[HTML]{d2ebd9}15.8 & \cellcolor[HTML]{d2ebd9}15 & \cellcolor[HTML]{d2ebd9}12 \\
\midrule
 & Claude Opus 4.6 & \cellcolor[HTML]{b5dec1}12.0 & \cellcolor[HTML]{7bc492}\textbf{27.5} & \cellcolor[HTML]{d2ebd9}19.3 & \cellcolor[HTML]{98d1aa}43 & \cellcolor[HTML]{7bc492}40 \\
\multirow{-2}{*}{{\tiny\bfseries\rotatebox[origin=c]{90}{\shortstack{Claude \\ Code}}}} & Claude Sonnet 4.6 & \cellcolor[HTML]{d2ebd9}7.8 & \cellcolor[HTML]{7bc492}21.5 & \cellcolor[HTML]{d2ebd9}20.4 & \cellcolor[HTML]{7bc492}\textbf{47} & \cellcolor[HTML]{7bc492}\textbf{47} \\
\midrule
 & Gemini 3.6 Flash & \cellcolor[HTML]{b5dec1}9.3 & \cellcolor[HTML]{7bc492}26.3 & \cellcolor[HTML]{d2ebd9}14.4 & \cellcolor[HTML]{98d1aa}36 & \cellcolor[HTML]{98d1aa}32 \\
\multirow{-2}{*}{{\tiny\bfseries\rotatebox[origin=c]{90}{\shortstack{Gemini \\ CLI}}}} & Gemini 3.1 Pro & \cellcolor[HTML]{eaf6ee}3.8 & \cellcolor[HTML]{b5dec1}15.5 & \cellcolor[HTML]{eaf6ee}5.1 & \cellcolor[HTML]{d2ebd9}11 & \cellcolor[HTML]{d2ebd9}15 \\
\midrule
\multicolumn{2}{@{}l}{$C_0$ (initial)} & 0.7 & 12.5 & 0.8 & 12 & 14 \\
\multicolumn{2}{@{}l}{$C_\emptyset$ (empty)} & 0.8 & 0.5 & 100.0 & 8 & 11 \\
\bottomrule
\end{tabular}
\captionof{table}{\textbf{Specifications and the plans they lead to.} \textbf{Left:} the specification each system submits, scored
against the acceptable plans ($\times100$). \textbf{Right:} the share of instances on which a fixed planner, given that
specification, delivers a plan the user accepts; Flash is Gemini 3.6 Flash, Luna is GPT-6 Luna. Within a harness,
models are ordered by $\mathrm{F}_{1,w}$. Shading: the value as a share of $C_{\mathrm{ref}}$'s.
Bold: best system per column.}
\label{tab:main-results}

\end{minipage}\hfill
\begin{minipage}[t]{\columnwidth}\vspace{0pt}
\centering
\footnotesize
\setlength{\tabcolsep}{1.0pt}
\renewcommand{\arraystretch}{1.3}
\newcolumntype{N}{>{\centering\arraybackslash}p{24.5pt}}   
\begin{tabular}{@{}cl NNN NN@{}}
\toprule
\multicolumn{2}{@{}l}{\textit{ReSHAPE Shopping}} & \multicolumn{3}{c}{Specification} & \multicolumn{2}{c}{Accepted (\%)} \\
\cmidrule(lr){3-5}\cmidrule(l){6-7}
& Model & $\mathrm{F}_{1,w}$ & $P_w$ & $R_w$ & Flash & Luna \\
\midrule
\multicolumn{2}{@{}l}{$C_{\mathrm{ref}}$ (reference)} & 26.5 & 22.8 & 52.8 & 38 & 57 \\
\midrule
 & Gemini 3.6 Flash & \cellcolor[HTML]{d2ebd9}9.1 & \cellcolor[HTML]{7bc492}20.3 & \cellcolor[HTML]{eaf6ee}10.5 & \cellcolor[HTML]{7bc492}36 & \cellcolor[HTML]{98d1aa}\textbf{41} \\
 & Kimi K3 & \cellcolor[HTML]{d2ebd9}8.9 & \cellcolor[HTML]{b5dec1}12.3 & \cellcolor[HTML]{d2ebd9}\textbf{13.4} & \cellcolor[HTML]{7bc492}36 & \cellcolor[HTML]{b5dec1}30 \\
\multirow{-3}{*}{{\scriptsize\bfseries\rotatebox[origin=c]{90}{\shortstack{Open \\ Hands}}}} & gpt-oss-120b & \cellcolor[HTML]{eaf6ee}5.2 & \cellcolor[HTML]{d2ebd9}5.8 & \cellcolor[HTML]{d2ebd9}11.2 & \cellcolor[HTML]{98d1aa}29 & \cellcolor[HTML]{b5dec1}24 \\
\midrule
 & gpt-oss-120b & \cellcolor[HTML]{d2ebd9}6.2 & \cellcolor[HTML]{b5dec1}11.8 & \cellcolor[HTML]{d2ebd9}12.2 & \cellcolor[HTML]{98d1aa}29 & \cellcolor[HTML]{b5dec1}28 \\
\multirow{-2}{*}{{\tiny\bfseries\rotatebox[origin=c]{90}{\shortstack{mini- \\ SWE- \\ agent}}}} & Gemini 3.6 Flash & \cellcolor[HTML]{d2ebd9}6.1 & \cellcolor[HTML]{98d1aa}17.0 & \cellcolor[HTML]{eaf6ee}8.2 & \cellcolor[HTML]{7bc492}32 & \cellcolor[HTML]{b5dec1}26 \\
\midrule
{\tiny\bfseries\rotatebox[origin=c]{90}{\shortstack{Claude \\ Code}}} & Claude Opus 4.6 & \cellcolor[HTML]{b5dec1}\textbf{11.2} & \cellcolor[HTML]{7bc492}\textbf{23.4} & \cellcolor[HTML]{d2ebd9}13.3 & \cellcolor[HTML]{7bc492}\textbf{47} & \cellcolor[HTML]{98d1aa}39 \\
\midrule
 & Gemini 3.6 Flash & \cellcolor[HTML]{d2ebd9}8.5 & \cellcolor[HTML]{7bc492}21.4 & \cellcolor[HTML]{d2ebd9}10.6 & \cellcolor[HTML]{7bc492}39 & \cellcolor[HTML]{98d1aa}\textbf{41} \\
\multirow{-2}{*}{{\tiny\bfseries\rotatebox[origin=c]{90}{\shortstack{Gemini \\ CLI}}}} & Gemini 3.1 Pro & \cellcolor[HTML]{d2ebd9}7.0 & \cellcolor[HTML]{98d1aa}16.0 & \cellcolor[HTML]{d2ebd9}11.6 & \cellcolor[HTML]{98d1aa}26 & \cellcolor[HTML]{98d1aa}37 \\
\midrule
\multicolumn{2}{@{}l}{$C_0$ (initial)} & 2.6 & 3.4 & 6.1 & 26 & 22 \\
\multicolumn{2}{@{}l}{$C_\emptyset$ (empty)} & 0.1 & 0.0 & 100.0 & 9 & 11 \\
\bottomrule
\end{tabular}
\captionof{table}{\textbf{The same on ReSHAPE Shopping}, for the eight systems run there. Columns, ordering, and shading are as in Table~\ref{tab:main-results}.}
\label{tab:main-results-shopping}

\vspace{10pt}
\centering
\footnotesize
\setlength{\tabcolsep}{1.0pt}
\newcolumntype{N}{>{\centering\arraybackslash}p{24.5pt}}
\begin{tabular}{@{}l NNN NN@{}}
\toprule
& \multicolumn{3}{c}{Specification} & \multicolumn{2}{c}{Accepted} \\
\cmidrule(lr){2-4}\cmidrule(l){5-6}
& $\mathrm{F}_{1,w}$ & $P_w$ & $R_w$ & Flash & Luna \\
\midrule
Spearman $\rho$ & .81 & .90 & .40 & .94 & .59 \\
Kendall $\tau_b$ & .64 & .79 & .21 & .87 & .40 \\
\bottomrule
\end{tabular}
\captionof{table}{\textbf{The two domains order the systems alike.} Rank correlation between the travel and the shopping
values of the 8 systems run in both (Tables~\ref{tab:main-results} and~\ref{tab:main-results-shopping}).}
\label{tab:cross-domain}

\end{minipage}
\end{table*}

\paragraph{\pipelinename.}
\label{sec:shapeforge}
Hand-building instances in which revision is both necessary and possible does not scale, so \pipelinename
converts an existing planning benchmark into such instances in three steps (Appendix~\ref{app:shapeforge}).
First, it adds attributes worth trading to every option, such as a flight's on-time rate.
Next, it finds a trade-off that the options actually offer, such as restaurants where a lower rating comes with much
cheaper meals.
Lastly, it builds a hidden user who would accept this trade-off but whose request does not state it.
The request becomes the initial specification $C_0=c_i\land c_k\land c_b$, whose condition $c_i$ on the sacrificed
attribute excludes the trade-off; $c_k$ is the deal-breaker and $c_b$ the stated budget.
Writing the trade-off into $C_0$ gives the \emph{reference specification} $C_{\mathrm{ref}}=[c_i\lor(c_i^{-}\land c_j)]\land c_k\land c_b$, which serves as one known-good revision and plays no part in defining $\mathcal{A}$.
\pipelinename keeps an instance only if the trade-off passes two checks.
It must be worth discovering: some plan that only $C_{\mathrm{ref}}$ admits beats the best plan under $C_0$ by at
least $\delta$.
It must also be needed: $\mathcal{S}(C_{\mathrm{ref}})$ matches $\mathcal{A}$ better than revisions that need no
knowledge of the trade-off, such as relaxing the budget.
No step uses agent performance.
Appendix~\ref{app:example} walks through one instance.

\paragraph{\benchname and ReSHAPE Shopping.}
\label{sec:travel-benchmark-composition}
\label{sec:shopping-benchmark}
We apply \pipelinename to the 102 medium and hard TravelPlanner queries \citep{xie-2024-travelplanner} included in
Collaborative Gym \citep{shao-2025-collaborative-gym} and to the 97 shopping requests of DeepPlanning
\citep{zhang-2026-deepplanning}, keeping the 88 and 87 for which it verifies a trade-off
(Appendices~\ref{app:travel-data} and~\ref{app:shopping}).
Since about half of the shopping requests state no budget, \pipelinename states one for each by a rule taken
from TravelPlanner's budgets.

\paragraph{Validating the instances and the simulated user.}
\label{sec:human-validation}
Since simulated users can diverge from real ones \citep{seshadri-2026-lost-in-simulation,zhou-2026-sim2real}, we ran a user study, which confirms that the trade-offs \pipelinename plants are ones people would take and that the simulated user judges them as people do.
In the study, 20 participants each judged six \benchname instances for a described traveler, choosing between a plan the initial
specification admits and a plan that only a revision of it admits (Appendix~\ref{app:human-study}).
The revision is either the instance's verified trade-off or a decoy that compensates with an attribute the traveler does not care about.
Under the verified trade-off, a majority of participants chose the plan outside the request on every instance; under the decoy, most kept the plan the request admits.
Their majority choice agrees with the simulated user in 11 of the 12 scenarios, with a tie in the last, and individual choices agree with it on 76\% (95\% CI [66, 85]), which is as high as any fixed answer can reach given how much participants disagree with one another.

\subsection{Agents and Evaluation}
\label{sec:agents-and-evaluation}

\paragraph{Agents.}
We pair four agent harnesses, mini-SWE-agent \citep{yang-2024-sweagent}, OpenHands \citep{wang-2025-openhands},
Claude Code \citep{anthropic-2025-claudecode}, and Gemini CLI \citep{google-2025-geminicli}, with six models into
fourteen systems (Appendix~\ref{app:agent-systems}), each requesting medium reasoning effort and otherwise keeping its
harness's defaults.

\paragraph{Task interface and runs.}
Each agent uses a command-line tool whose instruction (Appendix~\ref{app:prompt-agents}) states the objective and the available actions without suggesting a strategy: it may inspect options or check candidate plans for free, ask the user
a question, show the user one or two plans for feedback, and revise and submit the specification, in any order within the
budget $B=10$.
Each system runs once per instance against the simulated user of \S\ref{sec:simulated-user}, and a failed run is scored by its last valid specification (Appendix~\ref{app:run-statistics}).

\paragraph{Interaction protocols.}
To separate strategy from harness and model, we also run four fixed protocols with Gemini 3.6 Flash under the same
actions and budget, each restricted to one of the four kinds of interaction we find in the off-the-shelf logs
(\S\ref{sec:agents}).
Ask-before-Plan \citep{zhang-2024-ask-before-plan} asks preference questions before revising; Self-Directed Revision checks candidate plans itself without consulting the user, which we call self-checking; Exception Probing shows the user one plan outside the current search space at a time and revises after each judgment; and Trade-off Probing states a trade-off hypothesis and tests it with a pair of plans, at two units per pair (Appendix~\ref{app:protocols}).
We call the last two, which show plans outside the current specification, the probing protocols.

\paragraph{Planner and statistics.}
The fixed planner, Gemini 3.6 Flash with low thinking, searches the options for a plan that satisfies the
specification it is given (Appendix~\ref{app:planner}). For the main results we also report GPT-6 Luna as a second planner (Appendix~\ref{app:planner-robustness}).
Sources are compared by a paired bootstrap over instances with 1,000 resamples.

\paragraph{Specifications analyzed.}
Our analysis covers 6,325 specifications: on both benchmarks, the baselines $C_0$, the empty specification,
and $C_{\mathrm{ref}}$, and the final specifications of the agent systems, 14 on \benchname and 8 on ReSHAPE
Shopping; on \benchname, also those the four protocols hold after each unit of feedback.

\section{Analysis}
\label{sec:analysis}

\subsection{What Makes a Revision Useful for Planning?}
\label{sec:usefulness}

\paragraph{Reshaping the specification matters.}
On \benchname, with the planner fixed (Table~\ref{tab:main-results}), $C_0$ leads to an acceptable plan on only 12\% of instances,
and ignoring the request is worse: under the empty specification the planner almost always delivers a plan, but
only 8\% are acceptable.
The two fail for opposite reasons, $C_0$ admitting almost none of the acceptable plans and the empty
specification almost nothing else.
The single verified trade-off of $C_{\mathrm{ref}}$ raises acceptance to 55\%, and the off-the-shelf systems span
10--47\%: what the planner is given decides much of what it achieves.
On ReSHAPE Shopping (Table~\ref{tab:main-results-shopping}), $C_{\mathrm{ref}}$ again raises acceptance over $C_0$, from 26\% to 38\% with Gemini 3.6 Flash and from 22\% to 57\% with GPT-6 Luna, and the empty specification does worst.

\begin{figure}[t!]
\centering
\includegraphics[width=\linewidth]{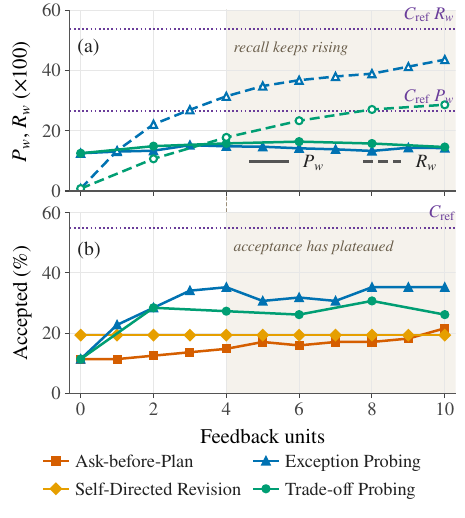}
\caption{\looseness=-1 \textbf{(a)} Precision (solid) and recall (dashed) of the probing protocols' specifications and \textbf{(b)} the acceptance rate of all four protocols, after each unit of feedback. Unit 0 is $C_0$, and Trade-off Probing spends two units per pair. In the shaded region past the fourth unit, recall keeps rising while acceptance has plateaued. Dotted lines mark $C_{\mathrm{ref}}$.}
\label{fig:quality-utility}
\end{figure}

\paragraph{A good revision first admits more plans and then stays selective.}
Figure~\ref{fig:quality-utility} gives the planner the specification each protocol holds after every unit of feedback.
The first revisions of both probing protocols raise recall from almost nothing, and acceptance rises with it: a single
judgment lets Exception Probing double the acceptance rate of $C_0$, and Trade-off Probing gains nearly all of its
improvement with its first pair.
Beyond the first few units (shaded in Figure~\ref{fig:quality-utility}), recall keeps growing while precision stays roughly flat, and acceptance no longer improves:
a planner delivers a single plan, so once the search space holds enough acceptable plans for it to find one, admitting
more adds little.
Precision then sets the level at which acceptance saturates: Exception Probing ends with a recall close to that of $C_{\mathrm{ref}}$ but half its precision, and is accepted 20 points less often; Self-Directed Revision has more than twice the recall of Ask-before-Plan but
half its precision, and is accepted less often (Table~\ref{tab:protocol-results}).

\paragraph{Search space quality predicts the planning outcome, and $\mathrm{F}_{1,w}$ tracks it as well as any measure we test.}
Instances differ in how hard they are to plan, so we compare specifications within an instance: for every pair of
specifications on the same instance of which the planner accepts exactly one, we ask whether a measure scores the
accepted one higher (the within-instance AUC, Appendix~\ref{app:measures}).
Over the fourteen systems and three baselines, $\mathrm{F}_{1,w}$ orders 87\% of such pairs correctly, against 85\%
for precision and 71\% for recall, and none of the variants we tried does reliably better; so search space quality can
stand in for running a planner.
On ReSHAPE Shopping, $\mathrm{F}_{1,w}$ orders 75\% of the pairs correctly with Gemini 3.6 Flash and 84\% with GPT-6 Luna.

\subsection{How Do Agents Revise, and How Should Feedback Be Spent?}
\label{sec:agents}

\paragraph{Current agents prioritize precision but cannot recover recall.}
Left free to interact, thirteen of the fourteen systems raise $\mathrm{F}_{1,w}$ significantly above $C_0$, but the best, mini-SWE-agent with
Kimi K3, reaches 14.0 against 21.5 for $C_{\mathrm{ref}}$ (Table~\ref{tab:main-results}).
The gap lies in recall: most systems reach a precision of 19--27, close to the 26.5 of $C_{\mathrm{ref}}$, but admit at
most a third of the acceptable plans, against more than half for $C_{\mathrm{ref}}$, so the plans the initial
specification excluded largely stay excluded.
Their effort goes to asking and self-checking, which together make up three quarters of their actions, while showing the user a plan makes up only a tenth: agents mostly judge candidate plans on the user's behalf instead of letting the user react to them.

Few systems write a trade-off at all: at most 39\% of any system's final specifications contain a meaningful one,
which relates two different attributes that both matter to the user, each restricted in the direction the user prefers
(Appendix~\ref{app:tradeoff-checker}).
ReSHAPE Shopping shows the same pattern: no system admits more than 13\% of the acceptable plans, against 53\% for $C_{\mathrm{ref}}$, though the most precise matches the precision of $C_{\mathrm{ref}}$, and asking and self-checking again make up most actions.

Across the eight systems run in both domains, the ordering by precision and by acceptance carries over from travel to shopping, but the ordering by recall does not (Table~\ref{tab:cross-domain}).

\begin{table}[t!]
\centering
\footnotesize
\setlength{\tabcolsep}{1.0pt}
\renewcommand{\arraystretch}{1.3}
\newcolumntype{N}{>{\centering\arraybackslash}p{24.5pt}}
\begin{tabular}{@{}l NNN N >{\centering\arraybackslash}p{28pt}@{}}
\toprule
& \multicolumn{3}{c}{Specification} & Acc. & Trade-off \\
\cmidrule(lr){2-4}
Protocol & $\mathrm{F}_{1,w}$ & $P_w$ & $R_w$ & (\%) & (\%) \\
\midrule
$C_{\mathrm{ref}}$ (reference) & 21.5 & 26.5 & 53.8 & 55 & 100 \\
\midrule
Ask-before-Plan & \cellcolor[HTML]{d2ebd9}5.3 & \cellcolor[HTML]{b5dec1}\textbf{14.7} & \cellcolor[HTML]{eaf6ee}9.6 & \cellcolor[HTML]{b5dec1}22 & \cellcolor[HTML]{eaf6ee}7 \\
Self-Directed Revision & \cellcolor[HTML]{eaf6ee}3.5 & \cellcolor[HTML]{d2ebd9}6.9 & \cellcolor[HTML]{b5dec1}22.9 & \cellcolor[HTML]{d2ebd9}19 & \cellcolor[HTML]{eaf6ee}3 \\
Exception Probing & \cellcolor[HTML]{98d1aa}\textbf{15.0} & \cellcolor[HTML]{b5dec1}14.2 & \cellcolor[HTML]{7bc492}\textbf{43.6} & \cellcolor[HTML]{98d1aa}\textbf{35} & \cellcolor[HTML]{eaf6ee}9 \\
Trade-off Probing & \cellcolor[HTML]{b5dec1}12.3 & \cellcolor[HTML]{b5dec1}14.6 & \cellcolor[HTML]{b5dec1}28.6 & \cellcolor[HTML]{b5dec1}26 & \cellcolor[HTML]{d2ebd9}\textbf{32} \\
\midrule
$C_0$ (initial) & 0.7 & 12.5 & 0.8 & 12 & 0 \\
\bottomrule
\end{tabular}
\caption{\textbf{Interaction protocols with Gemini 3.6 Flash} on 88 instances. Trade-off: runs ending with a meaningful
trade-off (Appendix~\ref{app:tradeoff-checker}). Shading as in Table~\ref{tab:main-results}. Bold: best per column.}
\label{tab:protocol-results}
\end{table}

\paragraph{Showing the user plans outside the current specification helps mitigate the imbalance.}
Across systems, the more plans an agent shows outside its current specification, the more acceptable plans its revision admits (Spearman's $\rho=.90$).
Since stronger models may both show more plans and revise better, we test this link under a fixed model, running each
kind of interaction on its own with Gemini 3.6 Flash (Table~\ref{tab:protocol-results}).
The two probing protocols reach $\mathrm{F}_{1,w}$ of 15.0 and 12.3, against 5.3 for Ask-before-Plan and 3.5 for
Self-Directed Revision, and lead the planner to acceptable plans more often, most of the gain arriving within the first
few units of feedback (Figure~\ref{fig:quality-utility}).
Asking barely moves the search space, and revising without the user admits mostly unacceptable plans, so
the user's reactions to concrete plans carry information the agent cannot recover on its own.
Exception Probing still gains recall at the cost of precision, so although it nearly doubles the $\mathrm{F}_{1,w}$ of
the same model off the shelf, its plans are accepted no more often (35\% vs.\ 31--36\%), as \S\ref{sec:usefulness}
predicts.
The same holds with Claude Sonnet 4.6 (Appendix~\ref{app:protocols-sonnet}): the two probing protocols reach
$\mathrm{F}_{1,w}$ of 13.8 and 12.1 against 2.6 and 5.1 for the controls and are accepted more often, while the
off-the-shelf Sonnet systems are accepted as often as its probing protocols.

\paragraph{Recognizing the user's trade-off is the bottleneck.}
Even when instructed to test trade-off hypotheses, the agent ends with a meaningful trade-off on only 28 of 88
instances; on those, Trade-off Probing is more precise than Exception Probing ($P_w$ 20.4 against 15.1) and nearly
matches $C_{\mathrm{ref}}$ ($\mathrm{F}_{1,w}$ 20.1 against 23.7), but on the other 60 it falls behind Exception Probing
(Table~\ref{tab:protocol-split}).
$C_{\mathrm{ref}}$ still outperforms every protocol and every system.

\section{Discussion and Conclusion}
\label{sec:conclusion}

\paragraph{Discussion.}\looseness=-1
Our results suggest three things beyond this benchmark.
First, evaluations of interactive planning should score the specification as well as the delivered plan: search space
quality needs no planner run, predicts the planner's acceptance, and keeps its orderings under a second planner
(Appendix~\ref{app:planner-robustness}) and in a second domain (Table~\ref{tab:cross-domain}), and unlike a delivered plan it shows whether a revision fails by admitting too
little or by keeping out too little.
Second, the order of work matters for agent design: early feedback is best spent showing plans outside the request, to
admit what it excluded, and later feedback on keeping out what the user would reject.
Agents that test trade-off hypotheses already revise most selectively, but only when they propose the right trade-off,
so proposing candidate trade-offs from the options actually on offer may be the most direct way to improve them.
Third, \pipelinename needs only options described by attributes and a space of legal plans; it already carries over from travel to shopping, and the same construction applies to other planning benchmarks, such as meal or course planning, whose requests likewise leave their trade-offs
unstated.

\paragraph{Conclusion.}

Planning agents inherit whatever specification they are given, and the one a user's request yields leaves out
plans the user would accept.
We made revising it measurable, by how well a revision's search space matches the plans the user would accept and by
what a planner finds within it, and analyzed it in travel and shopping planning.
Reshaping matters, and a good revision first admits more plans and then stays selective; current agents instead ask
about preferences and recover few of the plans the request excluded.
Planning agents should therefore treat a request as a hypothesis to test against the user's reactions to concrete
plans; recognizing which trade-off the user would accept remains the open problem.

\clearpage
\section*{Limitations}
\label{sec:limitations}

\paragraph{Users with fixed preferences.}\looseness=-1 
Our simulated user holds a fixed utility over plans, which makes acceptance exact and lets us score a revision over
every legal plan.
Real users may instead form or change their preferences as they see options, judge similar plans inconsistently, or
settle for a plan that is good enough \citep{simon-1955-behavioral-model}.
How revision should be measured, and which interaction works best, when the target itself moves with the interaction
is a question our setup cannot answer; it calls for users who revise their own requests.

\paragraph{One trade-off at a time.}\looseness=-1 
Each instance plants a single trade-off between two attributes of one option category.
Real requests can hide several trade-offs at once, including ones that span categories, such as a cheaper hotel that
pays for a better flight.
Whether the order we find, admitting first and selecting second, holds when trade-offs interact is open.

\paragraph{Revisions judged on one set of options.}
Search space quality is measured on the options of the instance, so a condition that happens to separate acceptable
from unacceptable plans there scores as well as the user's actual trade-off.
A revision meant to capture what the user wants should also hold on options the user has not seen; measuring this
transfer would separate understanding the user from fitting the instance.

\paragraph{Discovering, not presenting.}
We study how an agent discovers which plans the user would accept, with a user who judges every plan it is shown
exactly.
How an agent should present options so that a real user can reason over them, and how presentation interacts with
what the user discovers, is left to future work.

\ifanonymous\else
\section*{Acknowledgement}
This work is supported by the Okawa Foundation Research Grant, the Microsoft AI and New Future of Work Grant, Gemini Academic Program Award, and Amazon Research Award. We thank the user study participants for their valuable input.
\fi

\bibliography{custom}

\clearpage
\appendix
\raggedbottom
\makeatletter\setlength{\@fptop}{0pt}\setlength{\@fpsep}{14pt plus 2pt}\setlength{\@fpbot}{0pt plus 1fil}\setlength{\@dblfptop}{0pt}\setlength{\@dblfpsep}{14pt plus 2pt}\setlength{\@dblfpbot}{0pt plus 1fil}\makeatother
\renewcommand{\floatpagefraction}{0.6}\renewcommand{\dblfloatpagefraction}{0.6}

\FloatBarrier
\section{Related Settings}
\label{app:related-settings}

Table~\ref{tab:benchmark-comparison} compares \taskname with related planning and interaction settings (\S\ref{sec:related_works}).
It asks of each setting whether plans are composed over a real option set (grounded), whether the user reacts to
plans the agent shows (plan feedback), whether acceptance depends on a plan's overall utility rather than on each
constraint (trade-offs), and what it scores.
Most settings score the plan, or the set of options, that the agent finally delivers; GATE scores whether the
preferences it elicits predict the user's decisions on examples withheld from elicitation; \taskname scores the plans a
revised specification admits against the plans the user would accept, and what a planner finds among them.

\begin{table}[h!]
\centering
\footnotesize
\definecolor{cmpGreen}{HTML}{2E9E5B}
\definecolor{cmpYellow}{HTML}{DDA200}
\definecolor{cmpRed}{HTML}{D64B4B}
\newcommand{\cmpYes}{\makebox[1.6ex][c]{%
  \tikz[x=1.35ex,y=1.35ex,baseline=-0.03ex]{%
    \path[use as bounding box] (0,0) rectangle (1.10,1.04);
    \draw[draw=cmpGreen,line width=0.10em,line cap=round,line join=round]
      (0.10,0.47)--(0.36,0.14)--(1.02,0.99);}}}
\newcommand{\cmpPart}{\makebox[1.6ex][c]{%
  \tikz[x=1.35ex,y=1.35ex,baseline=-0.03ex]{%
    \path[use as bounding box] (0,0) rectangle (1.10,1.04);
    \draw[draw=cmpYellow,line width=0.10em,line cap=round,line join=round]
      (0.10,0.47)--(0.36,0.14)--(1.02,0.99);
    \draw[draw=cmpYellow,line width=0.10em,line cap=round]
      (0.38,0.63)--(1.06,0.63);}}}
\newcommand{\cmpNo}{\makebox[1.6ex][c]{%
  \tikz[x=1.35ex,y=1.35ex,baseline=-0.03ex]{%
    \path[use as bounding box] (0,0) rectangle (1.10,1.04);
    \draw[draw=cmpRed,line width=0.10em,line cap=round]
      (0.16,0.14)--(0.98,0.96)
      (0.16,0.96)--(0.98,0.14);}}}

\setlength{\tabcolsep}{2.5pt}
\renewcommand{\arraystretch}{1.05}
\newcommand{\cmpRow}[2]{\makecell[l]{#1\\{\scriptsize #2}}}
\resizebox{\linewidth}{!}{%
\begin{tabular}{@{}lcccl@{}}
\toprule
& \makecell[b]{Grounded\\planning} & \makecell[b]{Plan\\feedback} & \makecell[b]{Trade-\\offs} & Judged on \\
\midrule
\cmpRow{Ask-before-Plan}{\citep{zhang-2024-ask-before-plan}} & \cmpYes & \cmpNo & \cmpNo & delivered plan \\
\cmpRow{Flex-TravelPlanner}{\citep{oh-2025-flextravelplanner}} & \cmpYes & \cmpNo & \cmpPart & delivered plan \\
\cmpRow{UserBench}{\citep{qian-2025-userbench}} & \cmpPart & \cmpNo & \cmpNo & delivered options \\
\cmpRow{AdaPlanBench}{\citep{liu-2026-adaplanbench}} & \cmpNo & \cmpYes & \cmpNo & delivered plan \\
\cmpRow{GATE}{\citep{li-2025-eliciting-human-preferences}} & \cmpNo & \cmpPart & \cmpNo & held-out decisions \\
\midrule
\textbf{\taskname} & \cmpYes & \cmpYes & \cmpYes & \textbf{\makecell[l]{admitted plans +\\planner outcome}} \\
\bottomrule
\end{tabular}%
}
\caption{\textbf{Related settings}, by what each models and what it scores.}
\label{tab:benchmark-comparison}
\end{table}

\FloatBarrier
\section{\taskname on One Instance}
\label{app:example}

This appendix follows one instance of \benchname, travel:50, from the request to the planner's plan, so that the objects of \S\ref{sec:reshape} and the protocols of \S\ref{sec:agents} can be seen on real options.
Prices below are for the party of two.

\paragraph{Request and initial specification.}
The request has a budget of \$1,400 and reads:
\par\vspace{4pt}\noindent\begin{minipage}{\linewidth}
\userturn{Request}{Plan a 3-day trip for 2 people from Newark to Savannah from March 3, 2022 to March 5, 2022.}
\end{minipage}\par\vspace{4pt}
The initial specification $C_0$ holds three conditions: the chosen restaurants list at most two cuisines on average ($c_i$), the lodging is rated at least 3.0 ($c_k$, the deal-breaker), and the known total cost is at most \$1,400 ($c_b$).
The option pool holds 23 restaurants, and only three list two cuisines or fewer: New Raj Kachori Bhandar (\$24 a meal, rated 3.4), Country's Barbecue (\$118, 4.0), and Chutneez (\$104, 2.7).
Every plan $C_0$ admits therefore eats at these three; $C_0$ admits 20 legal plans, which differ only in lodging, transport, and attraction.

\paragraph{Hidden profile.}
Table~\ref{tab:example-profile} lists the user's six preferences.
Meal cost matters most to this user, and the number of cuisines a restaurant lists matters little.
The request nevertheless restricts the cuisine count and says nothing about meal cost.
This is the gap \pipelinename plants: the user would accept restaurants with more cuisines if meals are cheap.
The deal-breaker is the lodging rating, the more important of the two preferences outside the restaurant category, where the trade-off lies (Appendix~\ref{app:tolerance}).

\begin{table}[t]
\centering
\footnotesize
\setlength{\tabcolsep}{2.5pt}
\begin{tabular}{@{}lllll@{}}
\toprule
Attribute & Imp. & Expects & Prefers & Role \\
\midrule
Lodging rating & medium & medium & higher & $c_k$ \\
Attraction type & low & -- & historic site & \\
Meal cost & high & medium & lower & $j$ \\
Restaurant cuisines & low & -- & Italian & \\
Cuisine count & low & low & lower & $i$ \\
Restaurant rating & low & high & higher & \\
\bottomrule
\end{tabular}
\caption{\textbf{Hidden profile of travel:50.} Expects is the expectation of a numeric preference; Prefers is its direction or, for a categorical one, the liked value; the disliked values are markets and fast food. $c_k$ marks the deal-breaker, $i$ the sacrificed attribute, and $j$ the compensating one. The interaction style reads ``Explains requests in some detail and appreciates specific recommendations with brief context.''}
\label{tab:example-profile}
\end{table}

\paragraph{Acceptance and the reference specification.}
The best plan $C_0$ admits has utility $U(p_0^{*})=0.19$, so the threshold is $\tau=0.14$.
Of the 7,402,780 legal plans, 51,915 (0.7\%) are acceptable, and $C_0$ admits seven of them: its weighted precision is 58 and its weighted recall is near zero.
The reference specification relaxes the cuisine condition when meals are cheap.
$C_{\mathrm{ref}}$ requires a cuisine count of at most 2, or at most 6 with meals of at most \$56 on average, together with $c_k$ and $c_b$; eight restaurants have at most six cuisines and meals of at most \$56.
The reference plan drives both ways, stays in a 5.0-rated room, and eats at The Mad Teapot (\$46, rated 4.4, three cuisines), New Raj Kachori Bhandar, and Manohar Dairy (\$46, 4.4, three cuisines), for \$1,381 in all; its utility is 0.40, twice that of $p_0^{*}$.
$C_{\mathrm{ref}}$ admits 2,680 plans, 81\% of them acceptable, for a weighted precision of 94.6, a recall of 6.8, and an $\mathrm{F}_{1,w}$ of 12.8 (Table~\ref{tab:example-specs}).

\paragraph{The planner.}
Given $C_0$, Gemini 3.6 Flash found none of the 20 admitted plans in 14 calls and delivered nothing.
Given $C_{\mathrm{ref}}$, it delivered a plan that eats at Sr.\ Sol 1, Manohar Dairy, and The Mad Teapot and stays in the 3.0-rated Private Oasis room, which the user accepts.
Given the empty specification, it delivered a \$72-a-night room rated 1.0, which breaks the deal-breaker.

\begin{table}[t]
\centering
\scriptsize
\setlength{\tabcolsep}{2.5pt}
\begin{tabular}{@{}lrrrrrl@{}}
\toprule
Specification & $P_w$ & $R_w$ & $\mathrm{F}_{1,w}$ & Admitted & Accept. & Plan \\
\midrule
$C_\emptyset$ (empty) & 0.0 & 100.0 & 0.0 & 7,402,780 & 51,915 & rejected \\
$C_0$ (initial) & 58.1 & 0.0 & 0.0 & 20 & 7 & none \\
$C_{\mathrm{ref}}$ (reference) & 94.6 & 6.8 & 12.8 & 2,680 & 2,172 & accepted \\
\midrule
Ask-before-Plan & 58.1 & 0.0 & 0.0 & 20 & 7 & rejected \\
Self-Directed Revision & 19.8 & 3.6 & 6.0 & 6,560 & 1,646 & accepted \\
Exception Probing & 32.9 & 45.6 & 38.2 & 47,420 & 19,301 & accepted \\
Trade-off Probing & 56.2 & 18.0 & 27.3 & 14,420 & 7,784 & accepted \\
\bottomrule
\end{tabular}
\caption{\textbf{Specifications on travel:50.} Weighted precision, recall, and $\mathrm{F}_{1,w}$ ($\times100$); the number of legal plans each admits and how many of those the user accepts; and whether the plan Gemini 3.6 Flash delivers under it is accepted. Ask-before-Plan submitted $C_0$ unchanged, so its planner row is a second run on the same specification.}
\label{tab:example-specs}
\end{table}

\paragraph{The four protocols.}
Table~\ref{tab:example-specs} scores the specification each protocol submitted and what the planner made of it.
The excerpts below quote the simulated user's actual replies.

\emph{Ask-before-Plan} spent all ten units on questions and changed nothing after any answer, so it submitted $C_0$.
Two of its ten exchanges:
\par\vspace{4pt}\noindent\begin{minipage}{\linewidth}
\agentturn{Agent}{asks the preferred direction of the listed cuisine count.}
\userturn{User}{``I generally prefer places with a lower variety of cuisines.''}
\agentturn{Agent}{asks the preferred direction of transportation cost.}
\userturn{User}{``I don't have a strong preference either way on transportation costs. What matters most to me is finding something that works for my trip.''}
\end{minipage}\par\vspace{4pt}
No answer could say what the user would give up for cheaper meals, because no question of that form is allowed.
The planner delivered the three two-cuisine restaurants with the Private Oasis room, which the user rejects.

\emph{Self-Directed Revision} relaxed the cuisine count to three on its own, ``to admit budget-compliant plans.''
This admits 6,560 plans, 25\% of them acceptable, and the planner's plan is accepted.

\emph{Exception Probing} showed ten plans that the current specification excluded.
Its first three:
\par\vspace{4pt}\noindent\begin{minipage}{\linewidth}
\agentturn{Agent}{shows a plan that flies instead of driving, at \$1,536, with the three two-cuisine restaurants. \textbf{Hypothesis:} the budget is too strict when flying saves 18 hours.}
\userturn{User}{``I appreciate that there aren't too many cuisine options---that's actually what I'm looking for. But this plan isn't quite what I need overall.'' (rejected; no change)}
\agentturn{Agent}{shows a flight plan at \$1,396 whose restaurants list three cuisines on average, among them Dhaba On Wheels (\$36, rated 2.7) and Manohar Dairy. \textbf{Hypothesis:} the cuisine count may relax to three.}
\userturn{User}{``I'll go with this plan, though I have to say the restaurant ratings are a bit lower than what I'd normally look for. But let's see how it works out.'' (accepted; cuisine count relaxed to 3)}
\agentturn{Agent}{shows a \$1,406 flight plan at The Mad Teapot, New Raj Kachori Bhandar, and Manohar Dairy. \textbf{Hypothesis:} \$6 over budget is acceptable for better-rated restaurants.}
\userturn{User}{``I appreciate the suggestion, but the cuisine isn't really what I'm looking for. I'll go ahead with this plan though.'' (accepted; budget cutoff raised to \$1,406)}
\end{minipage}\par\vspace{4pt}
The reservation in the last reply concerns the kind of cuisine, not how many are listed: two of the three restaurants list fast food, which this user dislikes.
The remaining seven probes raised the budget cutoff step by step to \$1,538 and the cuisine count to four, and one probe of a 1.0-rated room was firmly rejected.
The agent never wrote a trade-off item; it loosened two cutoffs.
The result admits 47,420 plans, 41\% of them acceptable: recall rose from near zero to 45.6 at a precision of 32.9, the highest $\mathrm{F}_{1,w}$ of the four protocols and above $C_{\mathrm{ref}}$.
The planner's plan flies, eats at Sr.\ Sol 1, Country's Barbecue, and Manohar Dairy, costs \$1,514, and is accepted: the user tolerates an 8\% overspend for it.

\emph{Trade-off Probing} tested five hypotheses, each with a pair of plans.
Its first pair:
\par\vspace{4pt}\noindent\begin{minipage}{\linewidth}
\agentturn{Agent}{\textbf{Hypothesis:} the user accepts up to four cuisines on average when the restaurants are rated at least 4.0 on average. Plan A eats at Sr.\ Sol 1 (\$60, rated 4.6, four cuisines), Manohar Dairy, and The Mad Teapot; Plan B at Sr.\ Sol 1, Biryani Art (\$82, 3.3), and Hwealthcafe (\$50, 3.5). Both drive and stay at Private Oasis.}
\userturn{User on A}{``I appreciate the suggestion, but the cuisine isn't really what I'm looking for. I'll go ahead with this plan though.'' (accepted)}
\userturn{User on B}{``The restaurant cost works well for me, but I don't think this plan is right overall.'' (rejected)}
\end{minipage}\par\vspace{4pt}
The agent wrote the trade-off item: cuisine count at most 2, or at most 4 with restaurants rated at least 4.0.
It is meaningful by the rule of Appendix~\ref{app:tradeoff-checker}, since both attributes are active and restricted in the user's preferred direction, but it compensates with restaurant rating, a low-importance preference, where the user actually trades for meal cost.
Three later pairs tested the budget and raised its cutoff to \$1,600; one tested low lodging ratings and kept the deal-breaker.
The result keeps nearly the precision of $C_0$ but admits only 18.0 of the acceptable weight, and the planner's plan is accepted.

\paragraph{Off-the-shelf systems.}
Of the 14 systems, ten end with a specification under which the planner's plan is accepted.
The eight that spent seven or more of their ten units asking all submit specifications with a weighted recall under 4.
OpenHands with Kimi K3 asked twice and showed eight plans, and reaches a recall of 42.5.

\FloatBarrier
\section{Plans and Specifications}
\label{app:task}

This appendix details the objects of \S\ref{sec:task-formulation}: the legal plans a query allows and the language that
every specification is written in.

\subsection{Legal Plans}
\label{app:planning-domains}

For each query $q$, the legal plan space $\mathcal{P}$ contains every complete plan that satisfies the query's
immutable requirements.
In travel, a legal plan is an itinerary that follows the itinerary grammar.
A legal itinerary starts at the requested origin, visits every requested city exactly once in a contiguous block of
days, and returns to the origin.
Each movement between cities occupies a \emph{transition day}; all other days are \emph{stationary days} in the
current city (Table~\ref{tab:travel-plan-grammar}).

\begin{table}[t]
\centering
\footnotesize
\setlength{\tabcolsep}{4pt}
\begin{tabular}{@{}lcc@{}}
\toprule
Plan component & Transition day & Stationary day \\
\midrule
Transportation & Exactly one & None \\
Meals & None & Exactly three \\
Attractions & None & Exactly one \\
Accommodation & One, except final day & Exactly one \\
\bottomrule
\end{tabular}
\caption{\textbf{Daily itinerary grammar} defining the legal travel plans.}
\label{tab:travel-plan-grammar}
\end{table}

\looseness=-1 Every selected option must come from the query's option pool and match the corresponding route or city.
Flights must match the date of their transition day.
Restaurants and attractions cannot be reused within an itinerary, and accommodations must meet their minimum stays.

In shopping, a legal plan picks one product for every requested item from the products of that item's type and
target demographic; two requested items of the same type and demographic take two different products, and the order of
the products does not matter.

\subsection{Specification Language}
\label{app:constraint-language}

A specification has to be precise enough for a planner to check and for us to score exactly, yet rich enough to write
down a trade-off, so $C_0$, $C_{\mathrm{ref}}$, and every revision share one language.

\paragraph{Conditions.}
A condition, what planning benchmarks refer to as a constraint, is a rule that a plan must satisfy, such as ``restaurants rated at least 3.8''; a specification
is a list of them.
An atomic condition restricts one attribute of one option category.
A numeric condition bounds the attribute's mean over the options a plan selects in that category, in the attribute's
own unit; a meal-cost limit, for example, bounds the trip's average meal cost rather than each meal.
A categorical condition requires a value either of at least one selected option, as for cuisine, or of every selected option, as for room type.
A budget condition bounds the plan's known total cost: in travel, the full party's transportation, lodging, and meals,
since attraction prices are often unknown; in shopping, the products' prices.

\paragraph{Trade-offs.}
A specification is a conjunction of items, each an atomic condition or a trade-off item $c_i\lor(c_i^{-}\land c_j)$.
The relaxed condition $c_i^{-}$ uses the same numeric attribute, aggregation, and direction as $c_i$ with a strictly
weaker cutoff, and $c_j$ restricts a different attribute; a plan passes the item if it meets $c_i$, or meets both
$c_i^{-}$ and $c_j$.

\paragraph{Revisions.}
An agent revises a specification by adding or removing a condition, changing a numeric cutoff, replacing a categorical
value, or replacing the whole specification, which is how trade-off items are written.

\FloatBarrier
\section{Simulated User}
\label{app:simulated-user-model}

The simulated user realizes the three properties of acceptance in \S\ref{sec:task-formulation}: compensation, reference dependence, and deal-breakers.
A hidden profile states what the user cares about (\ref{app:hidden-user-model}), a utility scores complete plans
(\ref{app:utility}), an acceptance rule turns utility into accept or reject (\ref{app:tolerance}), and a language model
voices the resulting judgments without revealing the profile (\ref{app:simulated-user}).

\subsection{Hidden Profile}
\label{app:hidden-user-model}

A profile lists six active preferences, each on one attribute of the domain: travel has 18 attributes (Table~\ref{tab:travel-attribute-schema}), and shopping nine per product type, such as price, rating, brand, and fit.
A preference is numeric or categorical (Table~\ref{tab:hidden-profile-schema}).
Each carries an \texttt{importance}, which decides how much the attribute counts when preferences trade off.
A numeric preference also carries an \texttt{expectation}, how demanding the user's standard is, and, where the attribute can go either way, an \texttt{orientation}, whether lower or higher values are better.
For the traveler of Appendix~\ref{app:example}, for example, meal cost is a high-importance preference with a medium expectation, and the number of cuisines a restaurant lists a low-importance one.
The profile also carries a persona and an interaction style; these only shape how feedback is worded.

\pipelinename builds the profile around the instance's trade-off (Appendix~\ref{app:c0-construction}): two of the six
preferences are the sacrificed attribute $i$ and the compensating attribute $j$, and the other four are drawn at random,
with a fixed seed, from the attributes of the request's categories.

\begin{table}[t]
\centering
\footnotesize
\setlength{\tabcolsep}{6pt}
\renewcommand{\arraystretch}{1.12}
\begin{tabular}{@{}lll@{}}
\toprule
Type & Field & Values \\
\midrule
Numeric & \texttt{importance} & low, medium, high \\
 & \texttt{expectation} & low, medium, high \\
 & \texttt{orientation} & lower, higher (if both apply) \\
\midrule
Categorical & \texttt{importance} & low, medium, high \\
 & \texttt{liked} & one value of the attribute \\
 & \texttt{disliked} & another value of the attribute \\
\bottomrule
\end{tabular}
\caption{\textbf{Preference types} in the hidden profile.}
\label{tab:hidden-profile-schema}
\end{table}

\subsection{Utility}
\label{app:utility}

A plan's utility is computed in four steps.
\begin{enumerate}
\setlength{\itemsep}{1pt}
\item \textbf{Score each option.} For each active preference, a selected option scores how far above or below the
user's expectation it is. Table~\ref{tab:hidden-preference-scores} turns the option's value into a desirability
$s\in[0,1]$ and the expectation into a reference $r$; for a cost, the value is what the whole party pays, such as a
nightly price times the nights booked. The score is $\ell_j(o)=\operatorname{Round}(10\,[s_j(o)-r_j])$, an integer
from $-8$ to $8$, and zero for an option without a value.
\item \textbf{Combine within a category.} Each preference's scores are averaged over the options the plan selects in
the category, counting a lodging once per night, and the preferences are weighed by importance as shares of the category's total, with low, medium, and high weighing 1, 2, and 3.
\item \textbf{Combine across categories.} Each category is weighed by its share of the user's total importance, and the
sum is normalized:
\begin{equation}
U_{\mathrm{pref}}(p)=\frac{1}{Z}\sum_c\omega_c\,\frac{1}{|p_c|}\sum_{o\in p_c}\sum_j w_{cj}\,\ell_j(o),
\end{equation}
with $w_{cj}$ the weight of preference $j$ within category $c$, $\omega_c$ the weight of $c$, $p_c$ the options plan $p$
selects in $c$, and $Z=8$, the largest score, so that $U_{\mathrm{pref}}\in[-1,1]$.
\item \textbf{Subtract the budget penalty.} The request's budget $b$ is a soft preference, so moderate overspending can
be offset by strengths elsewhere. With $e=\max(0,\,\mathrm{cost}(p)/b-1)$ the share by which the plan's known total
cost exceeds $b$, the penalty $\pi(e)$ grows with it: overspending by a tenth costs almost nothing, by half 0.2, and
doubling the budget 1. The plan's utility is $U(p)=U_{\mathrm{pref}}(p)-\pi(e)$.
\end{enumerate}

\begin{table}[t]
\centering
\footnotesize
\setlength{\tabcolsep}{3.5pt}
\renewcommand{\arraystretch}{1.12}
\begin{tabular}{@{}lllll@{}}
\toprule
& & \multicolumn{3}{c}{Reference $r$} \\
\cmidrule(l){3-5}
Type & Desirability $s$ & low & med. & high \\
\midrule
Numeric & percentile among options & 0.25 & 0.50 & 0.75 \\
 & (reversed if lower is better) & & & \\
\midrule
Categorical & liked value: 1 & \multicolumn{3}{c}{0.50} \\
 & disliked value: 0 & & & \\
 & any other value: 0.5 & & & \\
\bottomrule
\end{tabular}
\caption{\looseness=-1 \textbf{Desirability and reference.} Desirability $s$ of an option's value, and the reference $r$: set by the profile's expectation for numeric preferences, else 0.5.}
\label{tab:hidden-preference-scores}
\end{table}

\subsection{Acceptance}
\label{app:tolerance}

\paragraph{Reference point.}
The user accepts nothing worse than what the request already secures, so the reference is the best plan the initial
specification admits, $p_0^{*}=\arg\max_{p\in\mathcal{S}(C_0)}U(p)$, found exactly over $\mathcal{S}(C_0)$.

\paragraph{Tolerance.}
Plans whose utilities differ by less than a just-noticeable difference are indistinguishable to the user
\citep{luce-1956-semiorders}, so the threshold sits a tolerance below the reference, $\tau=U(p_0^{*})-\delta$ with
$\delta=0.05$.

\paragraph{Deal-breaker.}
The deal-breaker is the condition $c_k$ of $C_0$ on the most important of the four other preferences that lies in
another category than the trade-off.
A plan that violates it is rejected whatever its utility.

\subsection{Interaction and Feedback}
\label{app:task-interface}
\label{app:simulated-user}

\paragraph{Actions.}
Inspecting options, checking plans, and editing the specification are free (\S\ref{sec:agents-and-evaluation}); a question about one attribute, its preferred direction or its importance, costs one unit, and each shown plan one unit.

A shown plan may lie outside the current specification; its judgment comes back with whether the specification admits
it and with its known cost against the budget.

\paragraph{Judgments.}
The hidden profile turns each shown plan into one of five response states (Table~\ref{tab:simulated-user-states}).
The state depends on the plan's margin $m=U(p)-\tau$ and on how much pulls against the verdict: $P$ and $N$ are the summed positive and negative contributions of its preferences and budget relative to the threshold.
A plan with $m>0$ is a strong accept if $N\le m$ and a soft accept otherwise.
A plan with $m=0$ is borderline.
A plan with $m<0$ is a soft reject if $P>-m$ and a strong reject otherwise.
A soft accept or borderline response may name one negative aspect, and a soft reject one positive aspect, but only when
a single attribute's contribution exceeds all others of the same sign combined; otherwise the response stays general.
A plan that violates the deal-breaker is a strong reject that names it.
A preference question returns the user's qualitative direction or liked value, an importance question whether the
attribute matters; numeric
expectations and cutoffs are never disclosed.

\begin{table}[t]
\centering
\footnotesize
\setlength{\tabcolsep}{4pt}
\begin{tabular}{@{}lll@{}}
\toprule
State & Verdict & May name \\
\midrule
Strong accept & accept & nothing \\
Soft accept & accept & one negative aspect \\
Borderline & accept & one negative aspect \\
Soft reject & reject & one positive aspect \\
Strong reject & reject & the deal-breaker, if violated \\
\bottomrule
\end{tabular}
\caption{\looseness=-1 \textbf{Response states} of the simulated user.}
\label{tab:simulated-user-states}
\end{table}

\paragraph{Wording.}
Claude Haiku 4.5 turns each judgment into one or two short sentences.
It sees only the response state, the aspect it may name, the plan's label, and the user's interaction style, never the
profile.
Its reply is checked so that it reveals no numbers, internal terms, instructions, or attributes beyond the permitted one.

\FloatBarrier
\section{Measuring a Revision}
\label{app:measuring}

\subsection{Exact Computation over the Legal Plans}
\label{app:region-computation}

The measures of \S\ref{sec:evaluation-metrics} sum plan weights over every legal plan.
A travel instance has a median of $1.2\times10^{13}$ legal plans and up to $3\times10^{23}$, and a shopping instance a median of $1.9\times10^{5}$ and up to $9.8\times10^{7}$. These are far too many to list one by one, so we count them instead, exactly and by one method for both domains.
The method rests on two facts: a plan's utility is a sum of per-category scores, and its known cost a sum of
per-category costs.
\begin{enumerate}
\setlength{\itemsep}{1pt}
\item \looseness=-1 For each plan structure, an itinerary skeleton in travel or a slot per item in shopping, and for each category, count the selections of options that give each pair of category score and category cost, keeping only selections that meet
the specification's conditions on it.
\item Combine the categories by convolution: selections scoring $a$ and $b$ in two categories add to $a+b$, and their
counts multiply. This gives the exact number of plans at each pair of total utility and total cost.
\item Subtract the budget penalty, which is linear in cost between a few breakpoints, and read off the number of
acceptable plans and the weights $\sum\lvert U(p)-\tau\rvert$ of the admitted and acceptable sets. A trade-off item, a
union of two branches, is counted by inclusion--exclusion.
\end{enumerate}

Acceptable plans make up a median of 0.06\% of legal plans in travel and 0.49\% in shopping, so a specification that
admits every plan reaches full recall at near-zero precision.

\subsection{Planner}
\label{app:planner}

We use two planners: Gemini 3.6 Flash for every result, and GPT-6 Luna for the main results (Appendix~\ref{app:planner-robustness}). Gemini 3.6 Flash runs at low thinking level, temperature 1.0, and at most 4,096 output tokens per call; GPT-6 Luna runs at low reasoning effort. Each receives the
query, the public option catalog, and one specification; it cannot ask the user or edit the specification.
It queries the catalog, checks candidate plans against the legality rules and the specification, and submits; the
first legal submission that satisfies the specification is its delivered plan.
An episode allows 50 model calls, five submissions, and 30 minutes.

A delivered plan is scored with the utility and acceptance rule of Appendix~\ref{app:simulated-user-model}.

\FloatBarrier
\section{\pipelinename, \benchname, and ReSHAPE Shopping}
\label{app:shapeforge}

\begin{figure*}[t!]
\centering
\includegraphics[width=\textwidth]{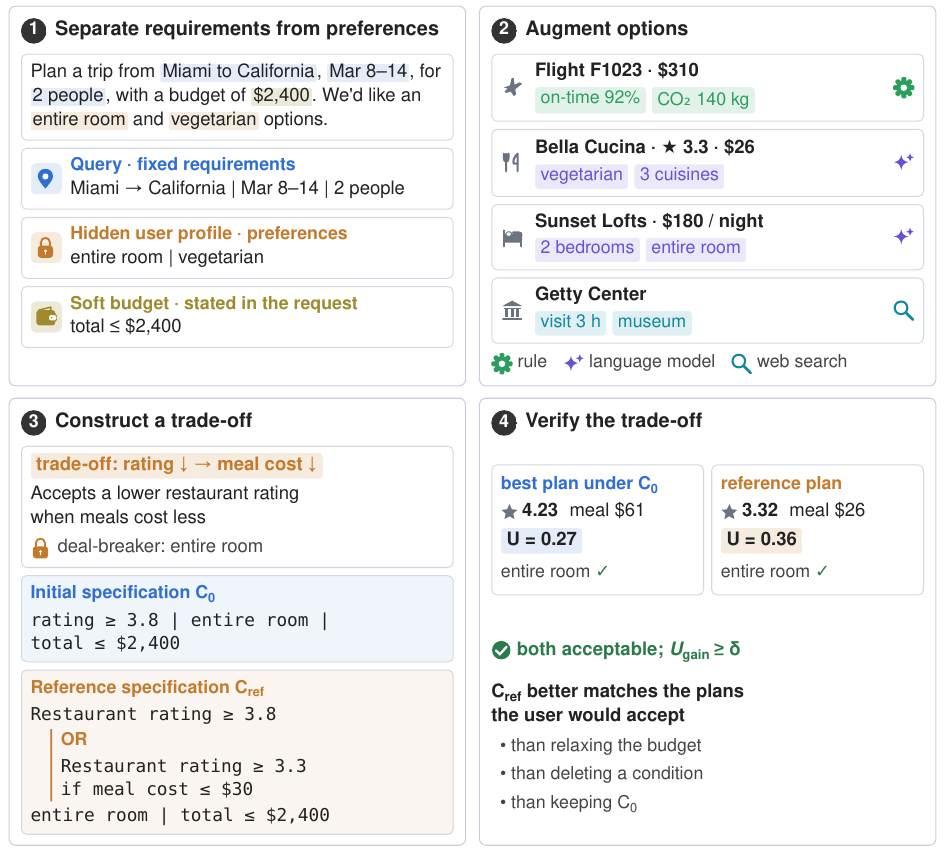}
\caption{The four steps of \pipelinename on the running example of Figure~\ref{fig:teaser}; step 2 illustrates the three
ways \pipelinename can add an attribute.}
\label{fig:shapeforge}
\end{figure*}

\pipelinename converts an existing planning benchmark into \taskname instances by adding what planning benchmarks
lack: attributes worth trading, a user who would accept a trade-off their request does not state, and a guarantee that
the trade-off matters (Figure~\ref{fig:shapeforge}).
The same steps build both domains; we describe them first and then each domain.

\subsection{Add Attributes Worth Trading}
\label{app:travel-attributes}
\label{app:attribute-augmentation}

A trade-off needs attributes that matter to a user and vary across options.
TravelPlanner describes each option by only a few, such as a restaurant's cuisine, rating, and average meal cost, so
\pipelinename adds eight, such as a flight's on-time rate or whether a restaurant offers vegetarian dishes, for 18 in all
(Table~\ref{tab:travel-attribute-schema}).
It derives a value by a rule over the option's fields that a language model helps write, by a language model reading the option's description, or by web
search, depending on where the information is (Table~\ref{tab:travel-generation-methods}).
DeepPlanning's products already carry enough, such as price, rating, monthly sales, review counts, brand, color, fit,
and season, so shopping needs no additions.

\begin{table}[t]
\centering
\footnotesize
\setlength{\tabcolsep}{5pt}
\renewcommand{\arraystretch}{1.08}
\begin{tabular}{@{}lll@{}}
\toprule
Category & Source & Added \\
\midrule
Transportation & cost & CO$_2$ \\
 & duration & on-time rate \\
 & departure time & \\
 & arrival time & \\
\midrule
Accommodation & nightly cost & bedroom count \\
 & review rating & property type \\
 & room type & \\
\midrule
Restaurants & meal cost & cuisine count \\
 & rating & dietary accommodation \\
 & cuisines & \\
\midrule
Attractions & & attraction type \\
 & & visit duration \\
\bottomrule
\end{tabular}
\caption{\textbf{Travel attributes} available for preferences and specifications.}
\label{tab:travel-attribute-schema}
\end{table}

\begin{table}[t]
\centering
\footnotesize
\setlength{\tabcolsep}{5pt}
\renewcommand{\arraystretch}{1.08}
\begin{tabular}{@{}lll@{}}
\toprule
Method & Model & Added attributes \\
\midrule
Rule & GPT-5.6 Sol & CO$_2$ \\
 & (writes the rules) & on-time rate \\
 & & bedroom count \\
 & & property type \\
 & & cuisine count \\
 & & dietary accommodation \\
\midrule
LLM & Claude Haiku 4.5 & none in travel \\
\midrule
Search & Gemini 3.6 Flash & attraction type \\
 & (search grounding) & visit duration \\
\bottomrule
\end{tabular}
\caption{\textbf{Attribute generation methods} supported by \pipelinename and their use in \benchname.}
\label{tab:travel-generation-methods}
\end{table}

\subsection{Build a User around an Unstated Trade-off}
\label{app:c0-construction}

A candidate trade-off pairs a numeric attribute $i$ that the user would sacrifice with a different attribute $j$ of
the same category that compensates, in one of three forms: paying more for a better attribute ($i$ is a cost), accepting
a worse attribute to pay less ($j$ is a cost), or exchanging one non-cost attribute for another.
\pipelinename tries every such pair.
For each, it builds a user who cares little about $i$ and much about $j$, with four other preferences drawn at random
with a fixed seed (Appendix~\ref{app:hidden-user-model}), and writes the two specifications:
\begin{itemize}
\setlength{\itemsep}{1pt}
\item $C_0=c_i\land c_k\land c_b$, what the user states up front: a threshold $c_i$ at the 20th, 40th, 60th, or 80th percentile of the request's values of $i$, the deal-breaker $c_k$ (Appendix~\ref{app:tolerance}), and the
stated budget $c_b$. Where the source request states no budget, \pipelinename states one by a rule calibrated to
TravelPlanner's own budgets.
\item $C_{\mathrm{ref}}=[c_i\lor(c_i^{-}\land c_j)]\land c_k\land c_b$, $C_0$ with the trade-off written in; the relaxed
threshold $c_i^{-}$ and the compensating condition $c_j$ range over the most lenient value, the reference plan's value,
and the midpoint between them.
\end{itemize}

\subsection{Verify That the Trade-off Matters}
\label{app:verification}

\pipelinename checks each candidate against the exact utility and
keeps it only if the trade-off is worth discovering and $C_{\mathrm{ref}}$ is the right revision rather than any
loosening:
\begin{itemize}
\setlength{\itemsep}{1pt}
\item $C_0$ admits at least one legal plan, so $p_0^{*}$ exists, and both stated conditions restrict the legal plans:
fewer plans meet $c_i\land c_k$ than meet $c_i$ or $c_k$ alone.
\item A reference plan that $C_{\mathrm{ref}}$ admits but $C_0$ excludes meets the deal-breaker and has
$U\ge U(p_0^{*})+\delta$.
\item Relative to $p_0^{*}$, the reference plan loses on $i$ and gains on $j$ more than it loses on $i$, and at least as
much as on any other preference, so the gain comes from the trade-off.
\item \looseness=-1 $C_{\mathrm{ref}}$ beats, in both $\mathrm{F}_{1,w}$ and unweighted $\mathrm{F}_1$, every revision that needs no
knowledge of the trade-off: $C_0$ itself, the budget alone, the best relaxation of the budget, and deleting a stated
condition.
\end{itemize}
Every instance thus holds a plan notably better than anything $C_0$ allows, reachable only through the planted
trade-off, and $C_{\mathrm{ref}}$ is one revision that reaches it.

\subsection{Select One Trade-off per Request}
\label{app:selection}

\looseness=-1 Most requests yield many candidates that pass.
\pipelinename keeps the one whose user accepts few of the legal plans, whose $C_0$ has low weighted precision, and whose
$C_{\mathrm{ref}}$ has the highest $\mathrm{F}_{1,w}$, with the first two bounded by the upper quartiles of \benchname, and
it balances the three forms across a dataset.

\subsection{\benchname}
\label{app:travel-data}
\label{app:instance-adjustment}

\benchname comes from the 102 medium and hard TravelPlanner queries in Collaborative Gym and follows the steps of
Appendices~\ref{app:attribute-augmentation}--\ref{app:selection} in an earlier version that differs in three places:
\begin{enumerate}
\setlength{\itemsep}{1pt}
\item \textbf{User.} The other four preferences came from one of 20 traveler profiles that Claude Opus 4.6 wrote
once, before seeing any query.
\item \textbf{Selection.} Candidates were tried in a fixed order, and the first to pass verification was kept.
\item \looseness=-1 \textbf{Adjustment.} An instance that failed verification was adjusted rather than discarded: by narrowing the
trade-off, trying another candidate, rescaling the stated budget, lowering the importance of uninvolved preferences,
shortening a seven-day trip to five days, or rescaling one category's prices. \end{enumerate}

Table~\ref{tab:benchmark-statistics} summarizes \benchname by trade-off form.

\begin{table}[t]
\centering
\footnotesize
\setlength{\tabcolsep}{3.2pt}
\begin{tabular}{@{}lrrrrr@{}}
\toprule
& & \multicolumn{2}{c}{$\mathrm{F}_{1,w}$} & \multicolumn{2}{c}{Utility} \\
\cmidrule(lr){3-4}\cmidrule(lr){5-6}
Trade-off & \# & $C_0$ & $C_{\mathrm{ref}}$ & $p_0^{*}$ & $p_{\mathrm{ref}}$ \\
\midrule
cost $\uparrow$ $\rightarrow$ attr.\ $\uparrow$ & 31 & 0.5 & 15.0 & 0.247 & 0.343 \\
attr.\ $\downarrow$ $\rightarrow$ cost $\downarrow$ & 27 & 1.2 & 31.2 & 0.306 & 0.398 \\
attr.\ $\downarrow$ $\rightarrow$ attr.$'$ $\uparrow$ & 30 & 0.4 & 19.3 & 0.279 & 0.387 \\
\midrule
All & 88 & 0.7 & 21.5 & 0.276 & 0.375 \\
\bottomrule
\end{tabular}
\caption{\textbf{\benchname by trade-off form}, each named as sacrificed $\rightarrow$ gained attribute, where attr.\ is a non-cost attribute. $\mathrm{F}_{1,w}$ is $\times 100$; utilities are means over instances. Instances offer on average 16.6 transportation options,
26.6 lodgings, 64.9 restaurants, and 40.4 attractions.}
\label{tab:benchmark-statistics}
\end{table}

\subsection{ReSHAPE Shopping}
\label{app:shopping}

ReSHAPE Shopping follows the steps of Appendices~\ref{app:attribute-augmentation}--\ref{app:selection}:
\begin{enumerate}
\setlength{\itemsep}{1pt}
\item \textbf{Source.} The 97 shopping requests of DeepPlanning \citep{zhang-2026-deepplanning}, each asking for a few
products of coarse types such as outerwear or footwear.
\item \textbf{Budget.} About half of the requests state no budget, so \pipelinename states one for every request
(Appendix~\ref{app:c0-construction}).
\item \looseness=-1 \textbf{Construction and selection.} 87 requests yield an instance: 29, 3, and 55 per form.
\end{enumerate}
Table~\ref{tab:shopping-construction} compares the instances with \benchname.
$C_0$ admits more of the acceptable plans in shopping, because a cart of two or three products leaves a stated
threshold fewer plans to exclude than a week of travel does, which is also why the planner succeeds under $C_0$ more often in shopping.

\begin{table}[t]
\centering
\footnotesize
\setlength{\tabcolsep}{3pt}
\begin{tabular}{@{}lcc@{}}
\toprule
& Travel & Shopping \\
\midrule
Instances & 88 & 87 \\
$P_w(C_0)$ & 0.37 & 1.76 \\
$R_w(C_0)$ & 0.03 & 3.14 \\
$P_w(C_{\mathrm{ref}})$ & 7.2 & 14.2 \\
$R_w(C_{\mathrm{ref}})$ & 51.2 & 51.7 \\
$\mathrm{F}_{1,w}(C_{\mathrm{ref}})$ & 9.3 & 22.6 \\
\midrule
Acc.\ $C_0$ (Flash) & 12 & 26 \\
Acc.\ $C_{\mathrm{ref}}$ (Flash) & 55 & 38 \\
Acc.\ $C_0$ (Luna) & 14 & 22 \\
Acc.\ $C_{\mathrm{ref}}$ (Luna) & 47 & 57 \\
\bottomrule
\end{tabular}
\caption{\textbf{Instances of \benchname and ReSHAPE Shopping.} Specification scores are medians over instances ($\times100$); acceptance (Acc.) is the share of
instances (\%).}
\label{tab:shopping-construction}
\end{table}

\FloatBarrier
\section{Experimental Details}
\label{app:experimental-details}

\subsection{Agent Systems}
\label{app:agent-systems}

\looseness=-1 All systems run through Harbor \citep{laude-2025-harbor} in Daytona sandboxes, with pinned harness versions.
Table~\ref{tab:agent-systems} lists the models of the systems in both domains.

\begin{table*}[t!]
\centering
\footnotesize
\setlength{\tabcolsep}{4pt}
\begin{tabular}{@{}llll@{}}
\toprule
Model & Identifier (provider) & Travel & Shopping \\
\midrule
Claude Sonnet 4.6 \citep{anthropic-2026-sonnet46} & \texttt{claude-sonnet-4-6} (Bedrock) & CC, MS, OH & -- \\
Claude Opus 4.6 \citep{anthropic-2026-opus46} & \texttt{claude-opus-4-6-v1} (Bedrock) & CC & CC \\
Gemini 3.6 Flash \citep{google-2026-gemini36flash} & \texttt{gemini-3.6-flash} (Vertex) & GC, MS, OH & GC, MS, OH \\
Gemini 3.1 Pro \citep{google-2026-gemini31pro} & \texttt{gemini-3.1-pro-preview} (Vertex) & GC, MS, OH & GC \\
gpt-oss-120b \citep{openai-2025-gptoss} & \texttt{gpt-oss-120b-1:0} (Bedrock) & MS, OH & MS, OH \\
Kimi K3 \citep{kimi-2026-k3} & \texttt{kimi-k3} (Bedrock) & MS, OH & OH \\
\bottomrule
\end{tabular}
\caption{\textbf{Models of the off-the-shelf systems} and the harnesses each runs in, per domain. CC: Claude Code; GC: Gemini CLI; MS: mini-SWE-agent; OH:
OpenHands.}
\label{tab:agent-systems}
\end{table*}

\subsection{Runs and Completion}
\label{app:run-statistics}

Besides the feedback budget, every run has a limit of 100 agent model calls and 30 minutes; calls to the
simulated user do not count against the call limit.
An infrastructure failure, such as a sandbox or provider error before the episode ends, is rerun, while the model's own
stopping, time-outs, and poor revisions are results.
The evaluated output is the last valid specification.
Table~\ref{tab:run-statistics} reports how each system's 88 runs ended on \benchname.

\begin{table}[t]
\centering
\footnotesize
\setlength{\tabcolsep}{2.2pt}
\begin{tabular}{@{}llrrrrr@{}}
\toprule
Harness & Model & Done & Early & Limit & Crash & Rerun \\
\midrule
Claude & Opus 4.6 & 88 & 0 & 0 & 0 & 45 \\
 \quad Code & Sonnet 4.6 & 87 & 0 & 1 & 0 & 20 \\
\midrule
Gemini & 3.6 Flash & 88 & 0 & 0 & 0 & 11 \\
 \quad CLI & 3.1 Pro & 88 & 0 & 0 & 0 & 10 \\
\midrule
mini-SWE- & Sonnet 4.6 & 88 & 0 & 0 & 0 & 18 \\
 \quad agent & 3.6 Flash & 77 & 11 & 0 & 0 & 11 \\
 & 3.1 Pro & 70 & 18 & 0 & 0 & 3 \\
 & Kimi K3 & 88 & 0 & 0 & 0 & 27 \\
 & gpt-oss-120b & 69 & 19 & 0 & 0 & 15 \\
\midrule
OpenHands & Sonnet 4.6 & 88 & 0 & 0 & 0 & 19 \\
 & 3.6 Flash & 88 & 0 & 0 & 0 & 11 \\
 & 3.1 Pro & 88 & 0 & 0 & 0 & 9 \\
 & Kimi K3 & 85 & 0 & 0 & 3 & 24 \\
 & gpt-oss-120b & 56 & 31 & 1 & 0 & 34 \\
\bottomrule
\end{tabular}
\caption{\textbf{How the runs ended} on \benchname, 88 per system. Done: the agent submitted a specification; Early: it ended the session
without submitting; Limit: it reached the call or time limit; Crash: the run failed after all reruns. Rerun: instances
that needed at least one rerun after an infrastructure failure. Runs that are not done are scored by their last valid
specification.}
\label{tab:run-statistics}
\end{table}

\subsection{Interaction Protocols}
\label{app:protocols}

The four protocols run only on \benchname and use the same simulated user, specification language, and feedback budget as the off-the-shelf systems; they differ only in the system prompt and the actions it allows. They run with Gemini 3.6 Flash at medium reasoning effort, temperature 0, and at most 4,096 output tokens per call.
After each unit of feedback the agent commits exactly one specification transaction.

Appendix~\ref{app:prompt-protocols} gives the prompts; in outline, they instruct the following.
\begin{itemize}
\setlength{\itemsep}{2pt}
\item \textbf{Ask-before-Plan.} Ask focused qualitative questions, each about one attribute and asking either its
preferred direction or its importance, at one unit each; never ask for an exact cutoff, a multi-attribute trade-off, or a
judgment of a concrete plan. The agent may stop asking early.
\item \textbf{Self-Directed Revision.} No feedback is available. List options and check complete plans for legality and
for whether the current specification admits them, which is free, and revise the specification on the user's behalf.
\item \looseness=-1 \textbf{Exception Probing.} In each round, show one complete legal plan that the current specification excludes
but that the agent judges worth recommending, stating which conditions exclude it, and revise after the judgment.
\item \looseness=-1 \textbf{Trade-off Probing.} In each round of two units, (1) identify an excluded candidate worth considering and
the condition that excludes it, (2) propose a tentative change, (3) state the trade-off hypothesis the change depends
on, (4) show a pair of plans designed to test it, recording predictions for all four outcomes before the judgments, and
(5) revise using the observed outcome, for at most five pairs.
\end{itemize}

\subsection{Interaction Protocols with Claude Sonnet 4.6}
\label{app:protocols-sonnet}
To check that the comparison of protocols does not depend on one model family, we also run the four protocols with
Claude Sonnet 4.6 on all 88 instances (Table~\ref{tab:protocol-sonnet}).
The prompts, actions, simulated user, feedback budget, and scoring are those of Appendix~\ref{app:protocols}; the
model runs at medium reasoning effort and temperature 1, with up to
20,000 output tokens per call.
The two probing protocols again lead both controls in $\mathrm{F}_{1,w}$ and in acceptance under both planners, and every $\mathrm{F}_{1,w}$ difference is significant under a paired bootstrap.
As with Gemini 3.6 Flash, probing raises $\mathrm{F}_{1,w}$ well above the off-the-shelf Sonnet systems but does not raise acceptance above theirs.

\begin{table}[t]
\centering
\scriptsize
\setlength{\tabcolsep}{2.5pt}
\renewcommand{\arraystretch}{1.15}
\begin{tabular}{@{}lrrrrrr@{}}
\toprule
& \multicolumn{3}{c}{Specification} & \multicolumn{2}{c}{Acc. (\%)} & Trade-off \\
\cmidrule(lr){2-4}\cmidrule(lr){5-6}
Protocol & $\mathrm{F}_{1,w}$ & $P_w$ & $R_w$ & Flash & Luna & (\%) \\
\midrule
\multicolumn{7}{@{}l}{\emph{Claude Sonnet 4.6}} \\
\quad Ask-before-Plan & 2.6 & 14.1 & 5.4 & 14 & 16 & 8 \\
\quad Self-Directed Revision & 5.1 & 10.1 & 30.1 & 19 & 22 & 5 \\
\quad Exception Probing & \textbf{13.8} & 17.2 & \textbf{40.4} & \textbf{39} & \textbf{36} & 27 \\
\quad Trade-off Probing & 12.1 & \textbf{18.5} & 28.2 & 32 & 32 & \textbf{34} \\
\midrule
\multicolumn{7}{@{}l}{\emph{Gemini 3.6 Flash (Table~\ref{tab:protocol-results})}} \\
\quad Ask-before-Plan & 5.3 & 14.7 & 9.6 & 22 & 22 & 7 \\
\quad Self-Directed Revision & 3.5 & 6.9 & 22.9 & 19 & 18 & 3 \\
\quad Exception Probing & \textbf{15.0} & 14.2 & \textbf{43.6} & \textbf{35} & 34 & 9 \\
\quad Trade-off Probing & 12.3 & 14.6 & 28.6 & 26 & \textbf{40} & \textbf{32} \\
\midrule
$C_{\mathrm{ref}}$ (reference) & 21.5 & 26.5 & 53.8 & 55 & 47 & 100 \\
$C_0$ (initial) & 0.7 & 12.5 & 0.8 & 12 & 14 & 0 \\
\bottomrule
\end{tabular}
\caption{\textbf{Interaction protocols with Claude Sonnet 4.6 and with Gemini 3.6 Flash} on 88 instances. Acc.: acceptance
rate under the Gemini 3.6 Flash and the GPT-6 Luna planner. Trade-off: runs ending with a meaningful trade-off
(Appendix~\ref{app:tradeoff-checker}). Bold: best protocol per column within a model.}
\label{tab:protocol-sonnet}
\end{table}

\subsection{Meaningful Trade-off Checker}
\label{app:tradeoff-checker}

\looseness=-1 A trade-off item $c_i\lor(c_i^{-}\land c_j)$ in a specification is meaningful when (1) attribute $i$ is active in the
user's profile and $c_i$ restricts it in the direction the user prefers, so that relaxing it to $c_i^{-}$ is a
concession; (2) attribute $j$ is active and $c_j$ restricts it in the preferred direction, or, for a categorical
attribute, includes a value the user likes; and (3) $i$ and $j$ are different attributes.
Attribute $i$ need not appear in $C_0$, and $i$ and $j$ need not share a category.

\FloatBarrier
\section{User Study}
\label{app:human-study}

The study asks whether people judge the planted trade-offs as the simulated user does (\S\ref{sec:human-validation}),
on instances of \benchname.

\subsection{Design}
\label{app:human-study-design}
\paragraph{Scenarios.}
We chose six instances whose trade-off can be read from a short profile, two per trade-off type
(Table~\ref{tab:human-study}).
For each we wrote two revisions of the initial specification.
The \emph{verified} revision is the instance's reference trade-off, and its two plans are the real $p_0^{*}$ and
reference plan, restricted to the stays or visits that differ.
The \emph{decoy} revision sacrifices the same attribute but compensates with an attribute the profile does not
weight; its outside plan is $p_0^{*}$ with only those two attributes changed, so under the additive utility the simulated
user prefers $p_0^{*}$.
In every scenario Plan A satisfies the initial specification and Plan B does not, so which plan lies outside the request
is never a cue to the condition.

\paragraph{Questions.}
Each scenario shows the traveler's profile, the initial specification, the revision with a one-line summary of what changed, and the two plans. The profile is the same in both versions and lists the attribute the decoy compensates with as unimportant.
Judging for the described traveler rather than for themselves, participants chose the plan the traveler would choose and rated from 1 to 5 how realistic the initial specification is, how plausible it is that a traveler would come to prefer the revision after seeing concrete options, and how confident they are.

\paragraph{Forms and participants.}\looseness=-1 
Two forms contain the same six scenarios in the same order with the conditions mirrored, so each participant sees three
verified revisions and three decoys, one of each per trade-off type.
We recruited 20 students at our institution through a department Slack channel; none were authors or had seen the
instances before, and none were compensated.
The study was approved by our institution's IRB, and participants gave informed consent before starting.
Participants were assigned to the two forms alternately and completed the study online in about 12--15 minutes.

\subsection{Analysis}
\label{app:human-study-analysis}
The analysis was fixed before data collection.
The primary test is a one-sided Wilcoxon signed-rank test of the within-participant difference in the share choosing Plan B under verified revisions and decoys: the mean difference is $.52$, with $p=.0001$.
Agreement with the simulated user is the share of choices matching its preference, with a 95\% bootstrap CI over
participants: 76\% $[66, 85]$ overall, 70\% for verified revisions and 82\% for decoys.
Because participants disagree among themselves, no fixed answer can exceed 76\% on these choices: choosing the
majority answer in every scenario reaches exactly that, and the simulated user does so, matching the majority in 11
scenarios and facing a 5--5 tie on the decoy of travel:43.

\begin{table}[t]
\centering
\footnotesize
\begin{tabular}{@{}llcc@{}}
\toprule
Instance & Trade-off type & Verified & Decoy \\
\midrule
travel:76 & pay more, better attribute & 7/10 & 1/10 \\
travel:93 & pay more, better attribute & 8/10 & 0/10 \\
travel:99 & worse attribute, pay less & 6/10 & 1/10 \\
travel:43 & worse attribute, pay less & 7/10 & 5/10 \\
travel:90 & one attribute for another & 8/10 & 1/10 \\
travel:45 & one attribute for another & 6/10 & 3/10 \\
\bottomrule
\end{tabular}
\caption{\looseness=-1 \textbf{User study choices.} Participants choosing the plan outside the initial specification (Plan B), by instance and revision. The simulated user picks Plan B for every verified revision and Plan A for every decoy.}
\label{tab:human-study}
\end{table}

\FloatBarrier
\section{Additional Results}
\label{app:additional-results}

\providecommand{\sig}{\rlap{$^\dagger$}}
\begin{table*}[t!]
\centering
\footnotesize
\setlength{\tabcolsep}{3.5pt}
\renewcommand{\arraystretch}{0.88}
\setlength{\aboverulesep}{0.2ex}
\setlength{\belowrulesep}{0.3ex}
\begin{tabular}{@{}llrrrrrrrr@{}}
\toprule
& & \multicolumn{4}{c}{Specification revision} & \multicolumn{4}{c}{Planning outcome} \\
\cmidrule(lr){3-6}\cmidrule(l){7-10}
Harness & Model & $\mathrm{F}_{1,w}$ & $P_w$ & $R_w$ & Done (\%) & Deliv.\ (\%) & Acc.\ (\%) & $\bar{U}$ & $\Delta U$ \\
\midrule
\multicolumn{2}{@{}l}{$C_{\mathrm{ref}}$ (reference)} & 21.5\sig & 26.5 & 53.8 & -- & 94 & 55 & 0.217 & +0.033\sig \\
\midrule
OpenHands & Claude Sonnet 4.6 & 9.5\sig & 23.5 & 22.6 & 100 & 90 & \textbf{47} & 0.210 & +0.031\sig \\
 & Gemini 3.1 Pro & 4.4\sig & 16.2 & 5.4 & 100 & 41 & 14 & 0.196 & +0.027 \\
 & Gemini 3.6 Flash & 8.7\sig & 24.5 & 14.7 & 100 & 81 & 34 & 0.196 & +0.022\sig \\
 & Kimi K3 & 12.2\sig & 19.2 & 31.5 & 97 & \textbf{92} & 36 & 0.218 & +0.030\sig \\
 & gpt-oss-120b & 2.1 & 6.0 & 13.1 & 64 & 55 & 10 & 0.183 & $-$0.008 \\
\midrule
mini-SWE-agent & Claude Sonnet 4.6 & 12.1\sig & 22.3 & 22.5 & 100 & 80 & 43 & 0.203 & +0.030 \\
 & Gemini 3.1 Pro & 7.3\sig & 20.8 & 13.2 & 80 & 60 & 25 & 0.214 & +0.021 \\
 & Gemini 3.6 Flash & 8.2\sig & 18.6 & 18.4 & 88 & 66 & 31 & 0.199 & +0.021\sig \\
 & Kimi K3 & \textbf{14.0}\sig & 25.9 & \textbf{31.6} & 100 & 85 & 44 & 0.206 & \textbf{+0.045}\sig \\
 & gpt-oss-120b & 2.9\sig & 4.8 & 15.8 & 78 & 47 & 15 & 0.181 & +0.003 \\
\midrule
Claude Code & Claude Opus 4.6 & 12.0\sig & \textbf{27.5} & 19.3 & 100 & 85 & 43 & \textbf{0.228} & +0.039\sig \\
 & Claude Sonnet 4.6 & 7.8\sig & 21.5 & 20.4 & 99 & \textbf{92} & \textbf{47} & 0.204 & +0.028\sig \\
\midrule
Gemini CLI & Gemini 3.1 Pro & 3.8\sig & 15.5 & 5.1 & 100 & 40 & 11 & 0.208 & +0.015 \\
 & Gemini 3.6 Flash & 9.3\sig & 26.3 & 14.4 & 100 & 76 & 36 & 0.203 & +0.006 \\
\midrule
\multicolumn{2}{@{}l}{$C_0$ (initial)} & 0.7 & 12.5 & 0.8 & -- & 66 & 12 & 0.156 & -- \\
\multicolumn{2}{@{}l}{$C_\emptyset$ (empty)} & 0.8 & 0.5 & 100.0 & -- & 98 & 8 & 0.091 & $-$0.086\sig \\
\bottomrule
\end{tabular}
\caption{\textbf{Full results of Table~\ref{tab:main-results}}, with completion and utility, on 88 instances with $B=10$.
Done: runs ending with the agent's own submission. Deliv./Acc.: instances where the fixed planner delivers a plan / an
acceptable plan; $\bar{U}$: mean utility of delivered plans; $\Delta U$: paired difference from $C_0$ on instances where both
deliver. $^\dagger$: the 95\% bootstrap CI of the difference from $C_0$ excludes zero. Bold: best system per column.}
\label{tab:main-results-full}
\end{table*}

These results back \S\ref{sec:analysis} on \benchname: which measure of search space quality tracks planning
(\ref{app:measures}), the full main results and the protocols split by trade-off outcome (\ref{app:region-change}),
and whether the outcomes hold under a second planner (\ref{app:planner-robustness}).

\subsection{Measures of Search Space Quality}
\label{app:measures}

Table~\ref{tab:measures} compares the measures by their within-instance AUC under both planners.

\begin{table}[t]
\centering
\footnotesize
\setlength{\tabcolsep}{5pt}
\begin{tabular}{@{}lcc@{}}
\toprule
& \multicolumn{2}{c}{Within-instance AUC [95\% CI]} \\
\cmidrule(lr){2-3}
Measure & Gemini 3.6 Flash & GPT-6 Luna \\
\midrule
$P_w$ & .85 [.81, .88] & .83 [.80, .86] \\
$R_w$ & .71 [.66, .75] & .71 [.67, .75] \\
$\mathrm{F}_{1,w}$ & .87 [.84, .90] & .85 [.82, .88] \\
$\mathrm{F}_{0.5,w}$ & .88 [.85, .91] & .85 [.82, .88] \\
$P_w \log R_w$ & \textbf{.89} [.86, .91] & \textbf{.85} [.82, .88] \\
\bottomrule
\end{tabular}
\caption{\textbf{How well each measure orders specifications by the planner's acceptance.} AUC: the share of same-instance pairs of specifications, one accepted and one not, that the measure orders correctly. Pairs come from the final specifications of the 14 systems and 3 baselines on 88 instances, and 95\% CIs from 1,000 bootstrap resamples of instances. $P_w \log R_w$ is $P_w(\log(R_w+\epsilon)-\log\epsilon)$ with $\epsilon=10^{-4}$.}
\label{tab:measures}
\end{table}

\begin{table}[t]
\centering
\footnotesize
\setlength{\tabcolsep}{4pt}
\begin{tabular}{@{}lrrr@{}}
\toprule
& $\mathrm{F}_{1,w}$ & $P_w$ & $R_w$ \\
\midrule
\multicolumn{4}{@{}l}{\emph{Trade-off Probing finds a trade-off} (28 instances)} \\
\quad Trade-off Probing & \textbf{20.1} & \textbf{20.4} & 42.6 \\
\quad Exception Probing & 18.2 & 15.1 & \textbf{57.3} \\
\quad $C_{\mathrm{ref}}$ & 23.7 & 29.3 & 55.3 \\
\multicolumn{4}{@{}l}{\emph{Trade-off Probing finds none} (60 instances)} \\
\quad Trade-off Probing & 8.6 & 11.9 & 22.1 \\
\quad Exception Probing & \textbf{13.5} & \textbf{13.9} & \textbf{37.3} \\
\quad $C_{\mathrm{ref}}$ & 20.4 & 25.2 & 53.1 \\
\bottomrule
\end{tabular}
\caption{\textbf{Protocols split by trade-off outcome}: Gemini 3.6 Flash on the instances where Trade-off Probing ends
with a meaningful trade-off and on the rest. Bold: best protocol per column and block.}
\label{tab:protocol-split}
\end{table}

\begin{table}[t]
\centering
\footnotesize
\setlength{\tabcolsep}{2.8pt}
\begin{tabular}{@{}lrrrrrr@{}}
\toprule
& \multicolumn{3}{c}{Gemini 3.6 Flash} & \multicolumn{3}{c}{GPT-6 Luna} \\
\cmidrule(lr){2-4}\cmidrule(l){5-7}
Specification & Deliv. & Acc. & $\bar{U}$ & Deliv. & Acc. & $\bar{U}$ \\
\midrule
$C_\emptyset$ (empty) & 98 & 8 & 0.091 & 98 & 11 & 0.055 \\
$C_0$ (initial) & 66 & 12 & 0.156 & 75 & 14 & 0.155 \\
Systems (mean) & 71 & 31 & 0.203 & 67 & 30 & 0.200 \\
$C_{\mathrm{ref}}$ (reference) & 94 & 55 & 0.217 & 89 & 47 & 0.226 \\
\bottomrule
\end{tabular}
\caption{\textbf{The planning outcome with two planners} on 88 instances. Deliv.\ and Acc.\ are in \%.}
\label{tab:planner-robustness}
\end{table}
Within-instance AUC asks, for every pair of specifications on the same instance of which the planner accepts exactly
one, whether a measure scores the accepted one higher, so that easy and hard instances do not mix.
Over 1,000 bootstrap resamples of instances, $\mathrm{F}_{1,w}$ orders such pairs reliably better than precision alone and than recall alone, while the two variants that weigh precision more, $\mathrm{F}_{0.5,w}$ and $P_w\log R_w$, do not reliably beat it.

Table~\ref{tab:acceptance-bins} shows how acceptance depends on both measures at once: it is lowest when both are low
and highest when both are high, and no row or column alone predicts it.

\begin{table}[t]
\centering
\footnotesize
\setlength{\tabcolsep}{4pt}
\begin{tabular}{@{}lrrrr@{}}
\toprule
& \multicolumn{4}{c}{$R_w$} \\
\cmidrule(l){2-5}
$P_w$ & 0--10 & 10--30 & 30--60 & 60--100 \\
\midrule
0--5 & 5 & 46 & 22 & 28 \\
5--15 & 4 & 39 & 54 & 67 \\
15--30 & 64 & 48 & 43 & 66 \\
30--100 & 58 & 81 & 77 & 92 \\
\bottomrule
\end{tabular}
\caption{\looseness=-1 \textbf{Acceptance by precision and recall.} Acceptance rate (\%) of the planner's delivered plans by the $P_w$ and $R_w$ of the specification, over all
5,368 specifications. Cells hold between 35 and 2,920 specifications.}
\label{tab:acceptance-bins}
\end{table}

\subsection{Planning Utility}
\label{app:region-change}

Table~\ref{tab:main-results-full} adds to Table~\ref{tab:main-results} the share of runs that end with the agent's own
submission (Done), the mean utility $\bar{U}$ of delivered plans, and the utility gain $\Delta U$ over $C_0$, the paired
difference in utility on instances where both deliver.
Table~\ref{tab:protocol-split} splits the protocol results by whether Trade-off Probing ends with a meaningful trade-off.

\subsection{Planner Robustness}
\label{app:planner-robustness}
To check that the planning outcome does not hinge on one planner, we give the specifications of the three baselines
and the fourteen systems also to GPT-6 Luna (Appendix~\ref{app:planner}).
Table~\ref{tab:planner-robustness} compares the two planners.
The orderings the paper relies on hold with both: the empty specification falls below $C_0$, the systems lie between
$C_0$ and $C_{\mathrm{ref}}$, in both acceptance and the utility of delivered plans, and within instances
$\mathrm{F}_{1,w}$ orders the specification each planner succeeds with above one it fails with about as often under both (Table~\ref{tab:measures}).
Table~\ref{tab:main-results-shopping} reports both for ReSHAPE Shopping.

\section{Prompts}
\label{app:prompts}

This appendix gives the instructions that the agents, the simulated user, and the planner receive (Prompts~\ref{prompt:agents}--\ref{prompt:planner}).
Where an instruction lists rules at length, we keep the rules that define the behavior and replace the rest with a bracketed note of what it says; the complete texts accompany the code.
Every protocol prompt ends with the schema and semantics of the specification language (Appendix~\ref{app:constraint-language}) and with the meaning of the five response states (Table~\ref{tab:simulated-user-states}); both are omitted.

\subsection{Off-the-Shelf Agents}
\label{app:prompt-agents}

Each off-the-shelf agent receives the task instruction of Prompt~\ref{prompt:agents} and reaches the environment through the command-line tool it describes.

\begin{prompt*}[p]
\refstepcounter{prompt}\label{prompt:agents}
\begin{prompttitle}[AgentBar]Prompt~\theprompt: Instruction to the off-the-shelf agents\end{prompttitle}
\begin{promptpar}[AgentBar]
Your task is to revise a travel-planning specification on behalf of a user.

The user has made a travel request and stated an initial specification: a set of conditions on attributes of the travel options. A specification determines which legal itineraries are admitted. The user's preferences are fixed but hidden from you. Your goal is to submit a revised specification that admits the itineraries this user would accept and excludes the ones they would not, across all legal itineraries, including itineraries never discussed with the user.

You may work in any way you choose. You can revise and submit the specification on your own, ask the user about their preferences, or show the user one or two itineraries and ask for feedback, in any order.
\end{promptpar}
\begin{promptsec}[AgentBar]
\textbf{Commands}

\begin{itemize}\setlength{\itemsep}{2pt}\setlength{\parskip}{0pt}
\item Inspect the task: reshape-task inspect, spec-schema, spec-current, itinerary-schema.

\item Read options: reshape-task options CATEGORY -{}-output FILE saves the full public catalog for one category.

\item Check itineraries: reshape-task check-plans FILE... reports whether each itinerary is legal and whether it satisfies the current specification. Free.

\item Ask a question: reshape-task ask ATTRIBUTE preference|importance. Costs 1 feedback unit. You may ask about the same attribute more than once.

\item Request feedback: reshape-task feedback FILE [FILE] shows the user one or two legal itineraries, which may lie outside the current specification. Costs 1 feedback unit per itinerary. Each response contains accepted, a response\_state, and the user's rationale.

\item Edit the specification: reshape-task spec add | remove | update-cutoff | replace-value | tradeoff | remove-tradeoff [arguments]. tradeoff replaces a condition with the original condition, OR a relaxed version of it together with a condition on another attribute. Append -{}-check to any edit to validate it without committing. Invalid edits change nothing and cost nothing.

\item Submit: reshape-task finish submits the current specification.
\end{itemize}
\end{promptsec}
\begin{promptsec}[AgentBar]
\textbf{Budget and limits}

\begin{itemize}\setlength{\itemsep}{2pt}\setlength{\parskip}{0pt}
\item You have 10 feedback units in total. Inspecting, checking, and editing are free, and you may keep editing after the feedback units are spent.

\item Each run is limited to 30 minutes and 100 model requests. If the run ends early, your latest valid specification is submitted.
\end{itemize}
\end{promptsec}
\begin{promptseclast}[AgentBar]
\textbf{Specification details}

\promptph{How the budget condition, the transport category, full-party prices, means over the selected options, and unknown values are defined.}

Use only the public task data.
\end{promptseclast}
\end{prompt*}

\subsection{Interaction Protocols}
\label{app:prompt-protocols}

Each protocol runs the same tool loop with its own system prompt: Ask-before-Plan (Prompt~\ref{prompt:ask}), Self-Directed Revision (Prompt~\ref{prompt:sdr}), Exception Probing (Prompt~\ref{prompt:ep}), and Trade-off Probing (Prompt~\ref{prompt:tp}).
Gemini 3.6 Flash and Claude Sonnet 4.6 receive the same text.
Self-Directed Revision is called Agent Decides in the code.

\begin{prompt*}[p]
\refstepcounter{prompt}\label{prompt:ask}
\begin{prompttitle}[AgentBar]Prompt~\theprompt: Ask-before-Plan\end{prompttitle}
\begin{promptpar}[AgentBar]
You are a travel-planning assistant using Ask-before-Plan. Revise the user's initial planning specification so that it better represents the plans the user would accept.

The query contains fixed planning requirements, such as destinations, dates, and party size. These requirements cannot be changed. The initial specification contains the user's starting editable conditions. You may inspect the query, available options, current specification, supported constraint language, and plan legality before deciding what to ask.

You have at most 10 user-feedback interactions.
\end{promptpar}
\begin{promptsec}[AgentBar]
\textbf{GATHER PREFERENCES}

You may ask the user focused qualitative questions without showing any concrete plan.

Each question must concern exactly one eligible preference attribute. You may ask about the same attribute repeatedly, including asking its preferred direction and then its importance. Each answer costs one feedback unit. Ask either: what direction the user prefers for that attribute; or how important that attribute is when choosing a trip.

Do not ask for an exact numerical cutoff. Do not ask the user to provide a complete preference profile. Do not ask a multi-attribute trade-off question. Do not present a concrete plan for judgment.

The public list of attributes is not a list of attributes the user necessarily cares about. The user may have no strong preference about an attribute.

After each answer, make exactly one constraint transaction before asking again. You may revise the current specification or use operations=[] for NO\_OP. \promptph{How a failed transaction is held pending until it is corrected.}
\end{promptsec}
\begin{promptseclast}[AgentBar]
\textbf{REVISE THE SPECIFICATION}

You may stop questioning as soon as you believe you have enough information. You do not need to use all 10 feedback interactions.

After an answer, you may add or remove supported conditions, update numeric cutoffs, replace supported categorical values, or leave the specification unchanged. Set finish=true on that transaction to stop questioning early; otherwise continue. At feedback unit 10, the transaction ends the session automatically. You may also finish before asking any question with a finish=true transaction.

Top-level items must hold together; a supported exception item uses A OR (A\_relaxed AND B). Do not invent unsupported expressions.
\end{promptseclast}
\end{prompt*}

\begin{prompt*}[p]
\refstepcounter{prompt}\label{prompt:sdr}
\begin{prompttitle}[AgentBar]Prompt~\theprompt: Self-Directed Revision\end{prompttitle}
\begin{promptpar}[AgentBar]
You are a travel-planning assistant using Agent Decides. Revise the user's initial planning specification so that it better represents the plans the user would accept.

\promptph{The fixed requirements and editable conditions, as in Ask-before-Plan.}

You have no user-feedback interactions. You cannot ask the user questions and cannot show the user plans. Decide on the user's behalf.
\end{promptpar}
\begin{promptseclast}[AgentBar]
\textbf{REVISE THE SPECIFICATION}

You may list options and check complete plans for legality and for whether the current specification admits them. These checks are free and do not involve the user.

When you judge that a different specification would serve the user better, commit a constraint transaction: add or remove supported conditions, update numeric cutoffs, replace supported categorical values, or leave the specification unchanged with operations=[]. You may commit several transactions. Set finish=true on a transaction to submit the specification and end the session. A failed transaction retains the last valid specification; correct it and commit again.

Top-level items must hold together; a supported exception item uses A OR (A\_relaxed AND B). Do not invent unsupported expressions.
\end{promptseclast}
\end{prompt*}

\begin{prompt*}[p]
\refstepcounter{prompt}\label{prompt:ep}
\begin{prompttitle}[AgentBar]Prompt~\theprompt: Exception Probing\end{prompttitle}
\begin{promptpar}[AgentBar]
You are a travel-planning assistant using Exception Probing. Revise the user's planning specification by testing concrete exceptions to its boundary.

The query contains fixed planning requirements that cannot be changed. The initial specification contains editable conditions; the current specification is the latest successfully saved set. A complete legal plan satisfies the fixed query requirements. An admitted plan also satisfies the current specification. Membership is separate from user acceptance.

You have 10 feedback units. Each independent single-plan judgment costs one unit.
\end{promptpar}
\begin{promptsec}[AgentBar]
\textbf{SELECT AND EXPLAIN A PROBE}

When any excluded legal plans exist, select one that you judge worth recommending from the task, public catalog, and user feedback so far, yet that the current specification excludes. Test whether the user accepts this exception. In the existing target\_hypothesis or accompanying explanation, briefly state why it is worth considering and which current condition or conditions exclude it. Do not choose an arbitrary excluded plan that you yourself find unappealing.

This recommendation judgment is yours, not hidden utility. Do not read the user profile, weights, or scores. You need not prove that the candidate is better than all admitted plans. Do not use a hidden ranking, utility tool, or second comparison plan. No sacrifice-for-benefit hypothesis is required; a single judgment cannot establish a compensating trade-off.

The excluded selection rule also applies when the current specification admits no legal plans. If every legal plan is admitted, instead select one admitted legal plan that you suspect should not be admitted, and state the possible admission error.

Submit exactly one new complete legal plan with a concise boundary-error hypothesis. Never repeat a judged plan. Give an accurate summary of the features that make it worth considering or suspect, including important drawbacks and the conditions responsible for any exclusion. Use accurate values and units; do not invent advantages or argue that the user should accept it.
\end{promptsec}
\begin{promptseclast}[AgentBar]
\textbf{USE FEEDBACK AND REVISE}

After every judgment, complete exactly one specification transaction, allowing NO\_OP. Treat the independent ACCEPT or REJECT as an observation about that plan, and the qualitative concern as additional evidence; unmentioned attributes may still matter.

Acceptance of an excluded plan reveals an exclusion error for that concrete plan. It does not establish why it was accepted or justify deleting the entire excluding condition. Rejection of an excluded plan is not an admission error, though its concern can provide further information. Rejection of an admitted plan reveals an admission error for that plan.

Use all public feedback so far, not just the latest ACCEPT. \promptph{How to weigh all feedback so far and report edits that change the status of judged plans.}

One acceptable exception can show that a condition is too strict without showing that its attribute should be unrestricted. Before deleting a condition, consider the enlarged region and evidence across different contexts; there is no required number of ACCEPTs.

When the feedback suggests that a condition is too strict only for plans with some other compensating feature, record that with an exception item rather than relaxing the condition for every plan: keep the original condition as A, set A\_relaxed to the weaker cutoff the evidence supports, and set B to the compensating condition. Relax or delete the condition outright only when the evidence suggests it is too strict whatever the other features.

You may retain, relax, tighten, replace, add, or delete supported conditions, including returning TRUE. Use an empty operations array for NO\_OP; a transaction need not change conditions. Use only the supplied compensation-exceptions/v1 constraint language.

Acceptance does not end Exception Probing. Continue from the latest successfully saved specification through 10 judgments, unless a runtime limit or error ends the session. Do not ask direct preference questions or submit plan pairs.
\end{promptseclast}
\end{prompt*}

\begin{prompt*}[p]
\refstepcounter{prompt}\label{prompt:tp}
\begin{prompttitle}[AgentBar]Prompt~\theprompt: Trade-off Probing\end{prompttitle}
\begin{promptlast}[AgentBar]
You are a travel-planning assistant using Tradeoff Probing. Repair the user's specification by finding compensating trade-offs that its hard conditions miss: a condition the user would relax for plans that are better in another respect.

The query fixes immutable planning requirements. A complete legal plan satisfies those requirements; an admitted plan also satisfies the current editable specification. Membership and user acceptance are separate. Use only the task, public catalog, current specification, and public feedback. You do not know hidden utility, weights, profiles, scores, or rankings. Do not assume preference directions for user-specific categories such as cuisine, room type, or attraction type.

A compensating trade-off is recorded as an exception item \{exception: \{A, A\_relaxed, B\}\}, which admits a plan if A holds, or if both A\_relaxed and B hold. A is the current condition on the sacrificed attribute, A\_relaxed is a weaker cutoff on the same attribute, and B is a condition on a different, compensating attribute.

You have 10 feedback units: two independent judgments per pair, costing 2, at most five pairs. Never repeat a judged plan. Each round follows five steps, recorded in a single propose\_plans call.

\promptph{The five response states and what each may disclose.}

1. \textbf{CHOOSE A CONDITION THAT MAY BE COMPENSATED}. Name one current condition (on the sacrificed attribute) that excludes legal plans which violate only that condition yet are clearly better on another attribute (the candidate compensating attribute). Give the catalog evidence: example values of both attributes in such plans. You may keep the same condition across rounds. \promptph{What to do when every legal plan is admitted.}

2. \textbf{WRITE THE EXCEPTION ITEM TO BE TESTED}. Write the exact exception item you would commit if the test supports it: A = the current condition, A\_relaxed = a specific weaker cutoff, B = a specific condition on the compensating attribute. Choose cutoffs that the pair in step 4 can test.

3. \textbf{STATE THE HYPOTHESIS}. State in one sentence that the user accepts the sacrificed attribute at the A\_relaxed level when B holds, but not without it. The alternatives are that the relaxation is acceptable without B, or unacceptable even with B.

4. \textbf{BUILD A CONTRAST PAIR AND MAP OUTCOMES TO EDITS}. Submit two distinct complete legal plans that both satisfy A\_relaxed but violate A, so both are currently excluded, with values of the sacrificed attribute as similar as practical. Plan A satisfies B; Plan B does not. Keep everything else as similar as practical, ideally differing in one option. \promptph{What to record about the pair.} State, in one sentence per branch, the edit that branch leads to under the rules of step 5. The design record is validated and saved before either judgment.

5. \textbf{REVISE FROM THE OBSERVED BRANCH}. After each judged pair, make exactly one transaction, allowing NO\_OP, and cite the observed branch in its note: ACCEPT\_REJECT (the compensation is needed): commit the exception item in place of condition A, with A\_relaxed no weaker than the sacrificed value the user accepted and B satisfied by Plan A but not by Plan B. ACCEPT\_ACCEPT (no compensation needed at this level): relax condition A to the tested level without a compensating condition. REJECT\_REJECT: NO\_OP; next round, test a milder relaxation or a stronger compensation, or move to another condition. REJECT\_ACCEPT (another factor is at work): NO\_OP; next round, try a different compensating attribute or another condition.

Once an exception item is committed, use later pairs to test its limits (a weaker A\_relaxed with B held, or a weaker B with A\_relaxed held) and update it when supported. After two rounds on the same condition in which Plan A is rejected, move to another condition. Acceptance judges the whole plan: do not relax a cutoff beyond values the user has accepted.
\end{promptlast}
\end{prompt*}

\subsection{Simulated User}
\label{app:prompt-user}

The hidden profile decides each judgment (Appendix~\ref{app:simulated-user}); Claude Haiku 4.5 only words it, with the system prompt of Prompt~\ref{prompt:user-system}.
These prompts call the agent ``the planner.''
For a shown plan, the request carries the plan's label, the response state, and the one aspect the reply may name, with the rules of Prompt~\ref{prompt:user-plan}.
For a question, the request carries the question and the authorized answer.
The rules ask the model to answer only the queried attribute, to express at most the one disclosed proposition, to volunteer no other preference, to give no numbers, thresholds, or repair advice, and to ask nothing back.
Every reply is then checked against the record before the agent sees it (Appendix~\ref{app:simulated-user}).

\begin{prompt*}[p]
\refstepcounter{prompt}\label{prompt:user-system}
\begin{prompttitle}[UserBar]Prompt~\theprompt: Simulated user: system prompt\end{prompttitle}
\begin{promptpar}[UserBar]
You are voicing one persistent traveler in an interaction with a travel-planning agent.

Your job is LANGUAGE REALIZATION ONLY.

The simulator has already decided what this traveler means. Express only that authorized meaning naturally in the traveler's interaction style.

You must not infer, select, strengthen, weaken, or add semantic content.
\end{promptpar}
\begin{promptseclast}[UserBar]
\textbf{INTERACTION STYLE:} [the profile's interaction style]

The interaction style controls only HOW you speak: concise vs. explanatory; direct vs. tentative; warm vs. matter-of-fact; conversational phrasing; sentence length and rhythm.

The interaction style is NOT evidence about WHAT the traveler prefers. Never infer a hidden preference, concern, reason, threshold, or factual claim from the interaction style.

The authoritative semantic record in the current request is the COMPLETE semantic content you may express. You may not add another preference or reason simply because it sounds plausible.

Speak as an ordinary traveler, not as an evaluator, benchmark, simulator, or optimization system.
\end{promptseclast}
\end{prompt*}

\begin{prompt*}[p]
\refstepcounter{prompt}\label{prompt:user-plan}
\begin{prompttitle}[UserBar]Prompt~\theprompt: Simulated user: wording a plan judgment\end{prompttitle}
\begin{promptpar}[UserBar]
Realize the following authoritative plan-feedback record: [label, response state, disclosed proposition or null]
\end{promptpar}
\begin{promptsec}[UserBar]
\textbf{Rules.} 1. Preserve the ACCEPT/REJECT verdict exactly. 2. Express at most the one disclosed proposition in the record. 3. If it is null, do not name or imply a specific hidden preference or attribute. 4. If a proposition is supplied, preserve its polarity and qualitative relation exactly. 5. Do not mention a second positive aspect, negative aspect, preference, or concern. 6. Do not infer anything from the plan beyond this record. 7. Never reveal numerical preference thresholds, utility values, weights, reference points, hidden scores, acceptance-region information, or simulator internals. 8. Do not tell the planner what constraint to add, remove, loosen, tighten, or set. 9. Do not ask the planner a question. 10. Prefer one or two short conversational sentences.
\end{promptsec}
\begin{promptseclast}[UserBar]
\textbf{Response-state meaning.} STRONG\_ACCEPT: clearly accept; do not introduce a reservation or attribute-specific reason. SOFT\_ACCEPT: accept while naturally expressing the supplied concern, if any. BORDERLINE: the verdict is ACCEPT; convey weak or borderline acceptance, using only the supplied negative proposition, if any. SOFT\_REJECT: reject while naturally acknowledging the supplied positive aspect, if any. STRONG\_REJECT: clearly reject; if a proposition is supplied, it is a firm requirement the plan violates: state it plainly as the reason. Otherwise do not invent an attribute-specific explanation.

Return STRICT JSON only: \{"text": "<natural traveler response>"\}
\end{promptseclast}
\end{prompt*}

\subsection{Planner}
\label{app:prompt-planner}

The planner (Appendix~\ref{app:planner}) receives the specification, the query, and catalog tools, with the instruction of Prompt~\ref{prompt:planner}.

\begin{prompt*}[p]
\refstepcounter{prompt}\label{prompt:planner}
\begin{prompttitle}[PlannerBar]Prompt~\theprompt: Planner\end{prompttitle}
\begin{promptlast}[PlannerBar]
You are a travel planning agent. Construct and submit one legal itinerary satisfying ONLY the supplied specification and immutable public task requirements. The supplied specification replaces any earlier specification. Do not invent option IDs. Query the public catalog yourself; no complete plans have been prepared for you. You cannot edit the specification or ask a user questions. No private preferences or scores are available.

Respond with JSON \{"actions": [\promptph{tool calls}]\} with at most 6 actions per response. Tools: catalog(category, filters, fields, offset, limit) queries public rows \promptph{filter syntax, pagination, and price units}. check\_plan(itinerary) checks a plan you construct against immutable legality rules and the supplied specification; returns public violations only. Numeric conditions apply to selection MEANS, not individually to every option. submit\_plan(itinerary): at most 5 formal submission attempts per episode, independently of the 50-model-request budget. The first legal plan satisfying the supplied spec is final and immediately ends planning. Five invalid submissions end with cannot\_find\_plan; this does not establish infeasibility. No private score is returned. Check before submitting when possible. stop(reason): end without a submitted plan if you cannot find one; this does NOT establish infeasibility.

Use the entire itinerary JSON schema. Transition days have transport, no meals/attractions, and an arrival-city hotel except the final return day. Stationary days require 3 distinct restaurants and 1 attraction; do not repeat restaurants/attractions across the trip. Respect hotel minimum nights. Build your own candidate and revise based on check\_plan feedback. Prefer efficient catalog queries; reserve requests to check and submit. Never claim submission without calling submit\_plan.
\end{promptlast}
\end{prompt*}

\end{document}